\documentclass{article} 
\usepackage{iclr2027_conference,times}

\usepackage{amsmath,amsfonts,bm}

\def\eqref#1{equation~\ref{#1}}

\def\1{\bm{1}}

\DeclareMathAlphabet{\mathsfit}{\encodingdefault}{\sfdefault}{m}{sl}
\SetMathAlphabet{\mathsfit}{bold}{\encodingdefault}{\sfdefault}{bx}{n}

\usepackage{url}

\usepackage{algorithm}
\usepackage{algpseudocode}
\usepackage{amsmath,amssymb}
\usepackage{bm}
\usepackage{graphicx}
\usepackage{subcaption}
\usepackage{wrapfig}

\usepackage{booktabs}
\usepackage{multirow}

\newtheorem{proposition}{Proposition}
\newtheorem{theorem}{Theorem}
\newtheorem{definition}{Definition}

\usepackage[hidelinks]{hyperref}
\usepackage{enumitem}

\title{Stable–Plastic Subspace Construction for Differentially Private Continual LoRA}

\author{Jin Liu \textsuperscript{1} \quad Ning Xi\textsuperscript{1}\quad Yinbin Miao\textsuperscript{1}\quad Junkang Liu\textsuperscript{2}\quad Ke Cheng\textsuperscript{1}\quad Jianfeng Ma\textsuperscript{1}\\
\\
\textsuperscript{1} Xidian University \quad
\textsuperscript{2} Tianjin University\\
}

\iclrfinalcopy 
\begin{document}

\maketitle

\begin{abstract}
Continual learning (CL) adapts pretrained models to a sequence of tasks while retaining knowledge acquired from earlier tasks. When task data are sensitive, differential privacy (DP) provides record-level protection, but maintaining both adaptation and retention under privacy constraints remains challenging. We study this problem in continual low-rank adaptation (LoRA) and observe that: 1) private training increases forgetting primarily during subsequent adaptation rather than initial task acquisition; 2) jointly perturbing both LoRA factors introduces a non-negligible bilinear noise interaction; and 3) DP aggravates the stability--plasticity dilemma, hindering both knowledge retention and adaptation to new tasks.
Based on these observations, we propose \textsf{DP-SPA}, which constructs a stable subspace from previously privatized updates and a plastic subspace from pretrained weights within its orthogonal complement. The method fixes these bases and privately optimizes only the LoRA factor, eliminating the bilinear noise interaction within each step. Before merging each task update, it rescales stable-direction coefficients using historical and current update energies. We establish a privacy guarantee and show that recalibration does not increase historical overlap and tightens a conditional upper bound on old-task loss increase. Experiments across five datasets, eight settings, and multiple privacy budgets show that \textsf{DP-SPA} improves average accuracy and reduces forgetting compared with the evaluated baselines.
\end{abstract}

\vspace{-2mm}
\section{Introduction}
Continual learning (CL, \cite{rolnick2019experience}) enables pretrained models to acquire new task knowledge sequentially while preserving previously learned capabilities. In many practical settings, however, continual adaptation is performed on sensitive task-specific data, exposing learned models to privacy attacks that can recover information about training data~\citep{fu2024membership, zhang2025soft, 317931}. Differential privacy (DP) addresses this concern by limiting how much the output distribution can change when a single record is added or removed~\citep{dwork2006differential}. Maintaining effective adaptation and knowledge retention under this constraint remains challenging.

Parameter-efficient fine-tuning (PEFT) reduces the number of trainable parameters for continual adaptation~\citep{liu2026parameter, qiao2024learn}. Among PEFT methods, Low-Rank Adaptation (LoRA, \cite{hu2022lora}) is a promising approach for CL by parameterizing each task update with two low-rank factors while keeping the pretrained backbone frozen. Existing continual LoRA methods, however, are largely designed for non-private training and often exploit task-dependent features, gradient information, or other data-derived statistics~\citep{lu2026continual, he2026task, qiu2026splitlora, liang2024inflora, wu2025sd}. 
Under DP, these data-dependent operations require appropriate privacy treatment and accounting. Meanwhile, fixed-factor LoRA has been explored in private federated learning~\citep{sun2024improving}, where fixing one factor avoids bilinear noise interaction. However, the magnitude of this interaction in DP continual learning remains insufficiently characterized, and fixed-factor optimization alone does not specify how to select adaptation directions that balance retention and adaptation. This raises a critical question: \emph{how should low-rank adaptation be structured to preserve historical knowledge while learning new tasks under DP?}

To answer this question, we first examine how DP affects continual LoRA and identify three key observations: 1) DP substantially exacerbates forgetting, with much of the degradation arising during subsequent private adaptation rather than initial task acquisition; 2) jointly privatizing the two LoRA factors introduces a non-negligible bilinear noise interaction; 3) DP worsens the stability--plasticity trade-off, leading to poorer knowledge retention at comparable levels of new-task adaptation.

These observations motivate \textsf{DP-SPA}, a \textbf{D}ifferentially \textbf{P}rivate \textbf{S}table-\textbf{P}lastic low-rank \textbf{A}daptation framework for CL. Building on fixed-factor optimization, \textsf{DP-SPA} addresses how to construct and update the adaptation basis across tasks under privacy constraints. It reconstructs a stable subspace from previously privatized updates and a plastic subspace from public pretrained weights within the stable subspace’s orthogonal complement. The resulting bases evolve across tasks but remain fixed within each task, while DP-SGD optimizes only the corresponding LoRA up-projection matrices. To reuse historical directions while controlling their modification, DP-SPA recalibrates stable-direction coefficients using historical and current update energies before merging. Basis construction and recalibration use only public information and privatized outputs, incurring no additional privacy cost. Theoretical analysis and extensive experiments further support the effectiveness of \textsf{DP-SPA}.

Overall, our contributions are threefold:
\begin{itemize}[leftmargin=1em, labelsep=0.4em]
\vspace{-2mm}
\item We characterize the effects of DP training on continual LoRA by separating task acquisition from forgetting during subsequent adaptation, quantifying the bilinear noise interaction under joint-factor optimization, and comparing knowledge retention at matched levels of new-task accuracy.

\item We propose \textsf{DP-SPA}, which constructs complementary stable and plastic subspaces from privatized history and public pretrained weights, fixes them within each task for single-factor private optimization, and recalibrates stable-direction updates using historical and current energies.

\item We establish the privacy guarantee of \textsf{DP-SPA}, characterize its stable--plastic geometry, and prove that recalibration reduces historical overlap and tightens a conditional bound on old-task loss increase. Extensive experiments across eight continual learning settings, five datasets, and multiple privacy budgets demonstrate higher average accuracy and lower forgetting than baselines.
\end{itemize}

\vspace{-2mm}
\section{Related Work}
\noindent
\textbf{Low-Rank Adaptation for Continual Learning.}
Recent work has explored LoRA for CL by controlling low-rank updates to mitigate task interference.
Representative approaches exploit gradient or
parameter subspaces~\citep{liang2024inflora,liu2025lora,wang2025plan},
gradient-space splitting~\citep{qiu2026splitlora}, residual-gradient
adaptation~\citep{luo2026keeplora}, magnitude--direction
decoupling~\citep{wu2025sd}, shared and task-specific
adapters~\citep{he2025cl}, and almost-orthogonal parameter
spaces~\citep{zhu2025bilora}. Continual LoRA
merging~\citep{qiao2026merge} further consolidates sequential
adaptations into a single adapter. 
These methods primarily address non-private CL, while components accessing private data require appropriate privacy treatment and accounting under DP. DP-SPA constructs stable and plastic subspaces from privatized updates and public pretrained weights, respectively, without accessing current private data to select bases.

\noindent
\textbf{Differentially Private Parameter-Efficient Fine-Tuning.}
Differentially private fine-tuning enables pre-trained models to adapt to sensitive downstream data with formal privacy guarantees. Parameter-efficient methods improve the privacy--utility trade-off by restricting private optimization to a small subset of parameters~\citep{yu2021differentially}, with subsequent work exploring bias-only adaptation~\citep{bu2024differentially}, low-rank adaptation for diffusion models~\citep{tsai2025differentially}, and private optimization within low-dimensional gradient subspaces~\citep{zheng2026differentially}.
In federated settings, FFA-LoRA, FedASK, and LA-LoRA explore different strategies for private LoRA optimization, including fixed-factor~\citep{sun2024improving}, sketch-based~\citep{wen2025differentially}, and alternating updates~\citep{liu2026rethinking}, respectively. These studies primarily consider individual downstream tasks rather than continual adaptation. Building on fixed-factor optimization, DP-SPA balances retention and adaptation through history-aware basis construction and update recalibration under DP.

\section{Understanding DP Low-Rank Continual Learning}

\subsection{Problem Formulation}
\label{subsec:problem_formulation}
We consider a sequence of $T$ tasks with private datasets
$\{\mathcal{D}^t\}_{t=1}^{T}$, where
$\mathcal{D}^t=\{(x_i^t,y_i^t)\}_{i=1}^{N^t}$ and
$y_i^t\in\mathcal{C}^t$.
The class spaces are disjoint across tasks, i.e.,
$\mathcal{C}^t\cap\mathcal{C}^{t'}=\emptyset$ for $t\neq t'$. In the class-incremental setting, after learning task $t$, the model predicts over all classes seen so far,
$\mathcal{C}^{1:t}=\bigcup_{\tau=1}^{t}\mathcal{C}^\tau$,
without access to task identity. Given a pretrained weight matrix $\boldsymbol W^0\in\mathbb R^{m\times n}$, LoRA~\citep{hu2022lora} represents the task-$t$ low-rank adaptation by $\boldsymbol B^t\boldsymbol A^t$, where $\boldsymbol A^t\in\mathbb R^{r\times n}$ and $\boldsymbol B^t\in\mathbb R^{m\times r}$ with $r\ll\min(m,n)$. In continual adaptation, each task $t$ starts from the model obtained after learning the preceding tasks and introduces an additive low-rank update: $\boldsymbol W^t = \boldsymbol W^{t-1} + \Delta\boldsymbol W^t, \Delta\boldsymbol W^t = \boldsymbol B^t\boldsymbol A^t$.
During task $t$, the previously learned weights $\boldsymbol W^{t-1}$ are kept frozen, while both low-rank factors $\boldsymbol A^t$ and $\boldsymbol B^t$ are trainable. We initialize $\boldsymbol A^t$ from a Gaussian distribution and set $\boldsymbol B^t=\boldsymbol 0$. We optimize the trainable LoRA parameters using DP-SGD~\citep{abadi2016deep}. For each sample $z_i^t=(x_i^t,y_i^t)\in\mathcal B^t$, where $\mathcal B^t\subseteq\mathcal D^t$ denotes a mini-batch, the per-sample gradient is clipped as ${\boldsymbol{g}}_i^t
    =
    {\nabla_{\boldsymbol{\phi}^t}
    \ell(\boldsymbol{\phi}^t;z_i^t)}
    /{\max\!\left(
    1,
    \|\nabla_{\boldsymbol{\phi}^t}
    \ell(\boldsymbol{\phi}^t;z_i^t)\|_2/C
    \right)}$, where $\boldsymbol\phi^t$ denotes the trainable LoRA parameters at task $t$ and $\ell$ denotes the per-sample loss. The privatized mini-batch gradient is $\widetilde{\boldsymbol{g}}^t=
((\sum_{i\in\mathcal{B}^t}\boldsymbol{g}_i^t+\mathcal{N}(\boldsymbol{0},C^2\sigma^2\boldsymbol{I}))/
|\mathcal{B}^t|)$, where $C$ is the clipping norm and $\sigma$ is the Gaussian noise multiplier calibrated to the target privacy budget. Each protected record belongs to exactly one task dataset, and each task satisfies $(\epsilon,\delta)$-DP. Theorem~\ref{thm:dp_main} formally states the overall privacy guarantee.

\begin{definition}[$(\epsilon, \delta)$-DP]
Let $\epsilon>0$ and $0<\delta<1$. A random mechanism $\mathcal{M}:\mathcal{X}^n\!\to\!\mathcal{O}$ satisfies $(\epsilon, \delta)$-DP if, for any two neighboring datasets $\mathcal{D}$ and $\mathcal{D}'$ that differ by the addition or removal of one record, and for any measurable $\mathcal{U}\subseteq\mathcal{O}$,
$\Pr[\mathcal{M}(\mathcal{D}) \in \mathcal{U}] \leq e^\epsilon \cdot \Pr[\mathcal{M}(\mathcal{D}') \in \mathcal{U}] + \delta.
$
\end{definition}

\subsection{Challenges of DP Low-Rank Continual Learning}
\label{subsec:challenges}
We consider a 20-task class-incremental sequence on
ImageNet-A~\citep{hendrycks2021natural} and fine-tune a ViT-B/16 model~\citep{dosovitskiy2021an}. By default, both LoRA factors $\boldsymbol{A}$ and $\boldsymbol{B}$ are jointly updated, with DP noise added to the gradients of both factors.

\begin{figure*}[h]
    \centering

    \begin{subfigure}[t]{0.32\textwidth}
        \centering \includegraphics[width=\linewidth]{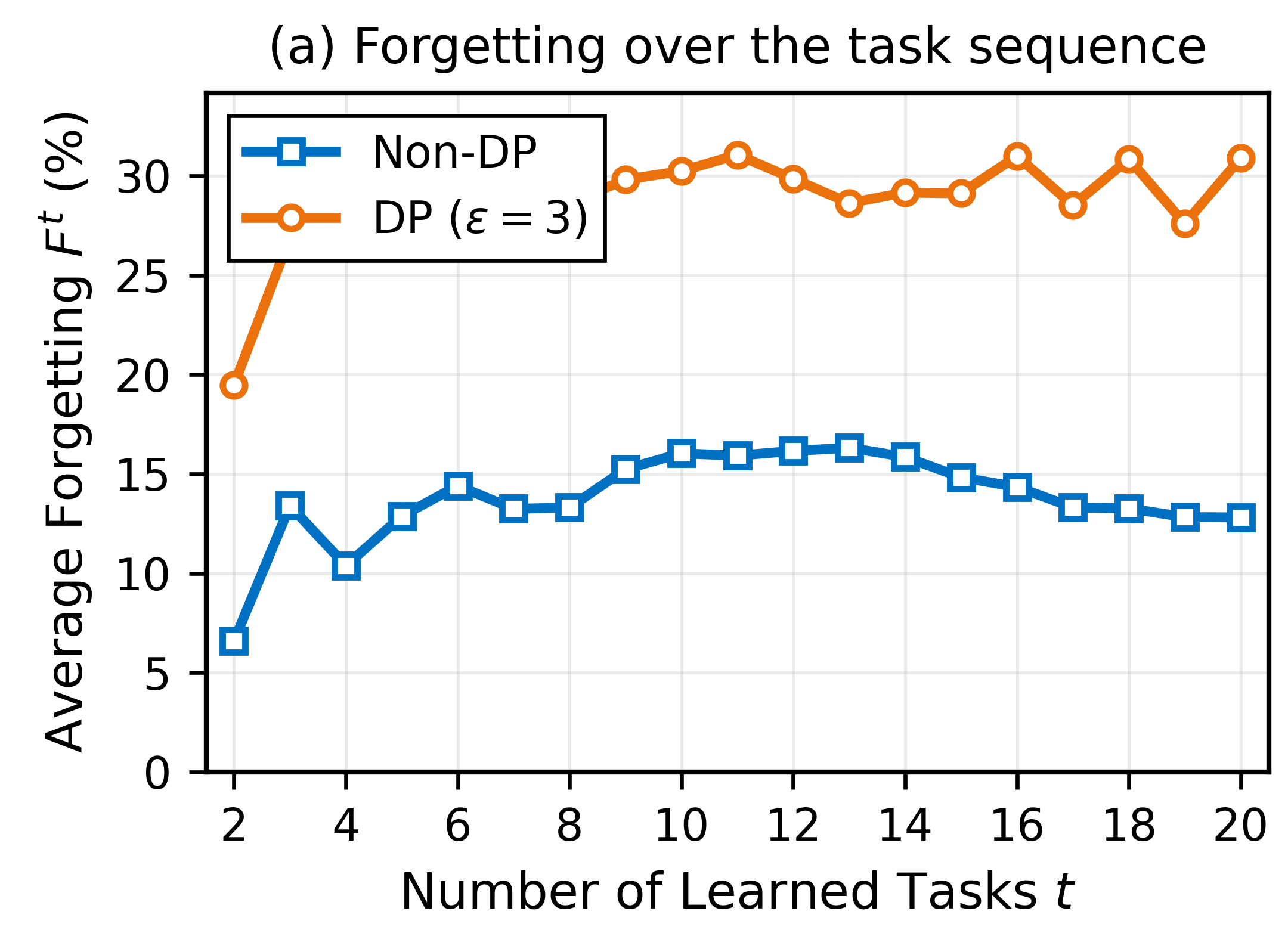}
        \label{fig:challenge1_a}
    \end{subfigure}
    \hfill
    \begin{subfigure}[t]{0.32\textwidth}
        \centering
\includegraphics[width=\linewidth]{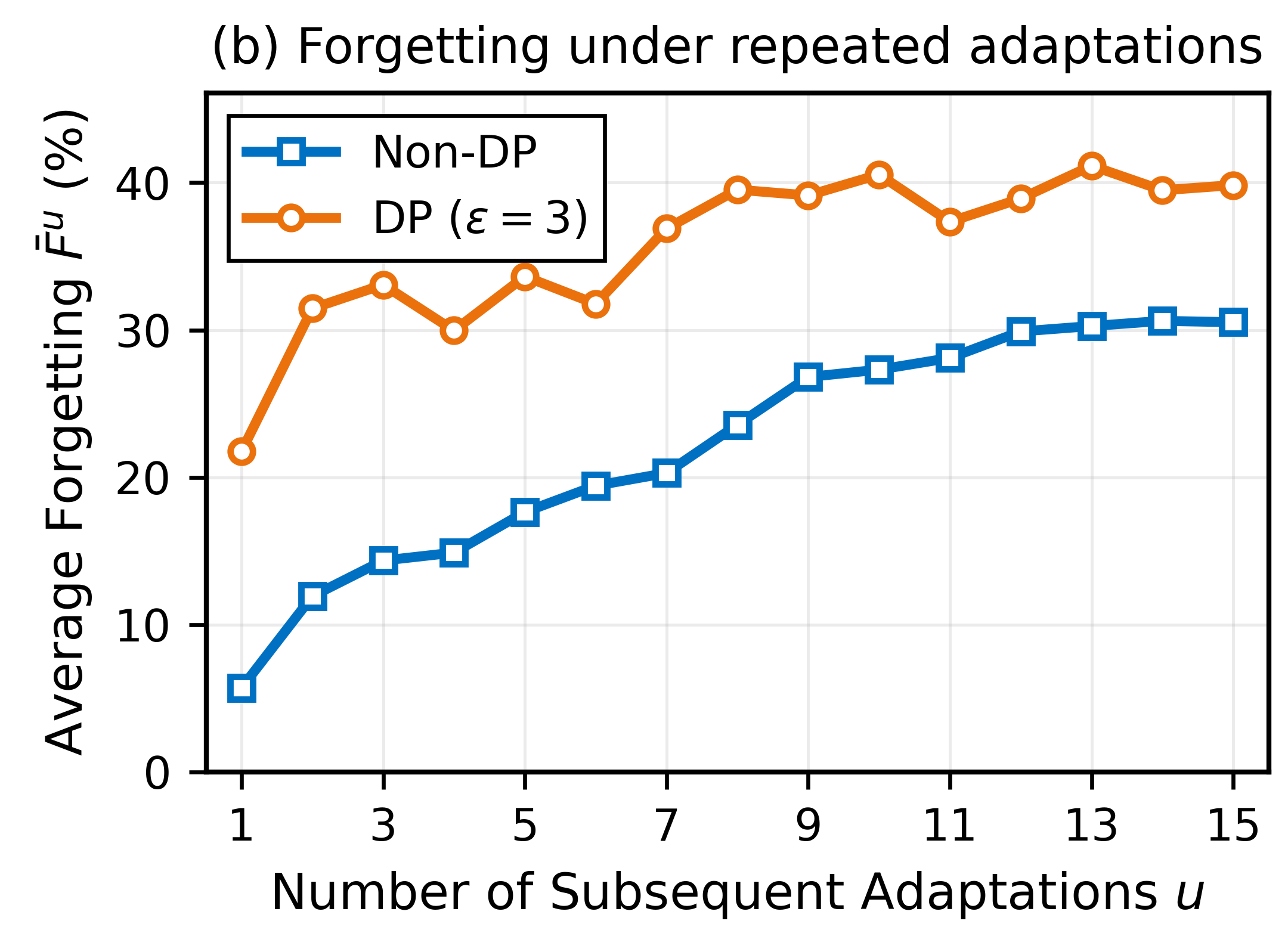}
        \label{fig:challenge1_b}
    \end{subfigure}
    \hfill
    \begin{subfigure}[t]{0.32\textwidth}
        \centering
\includegraphics[width=\linewidth]{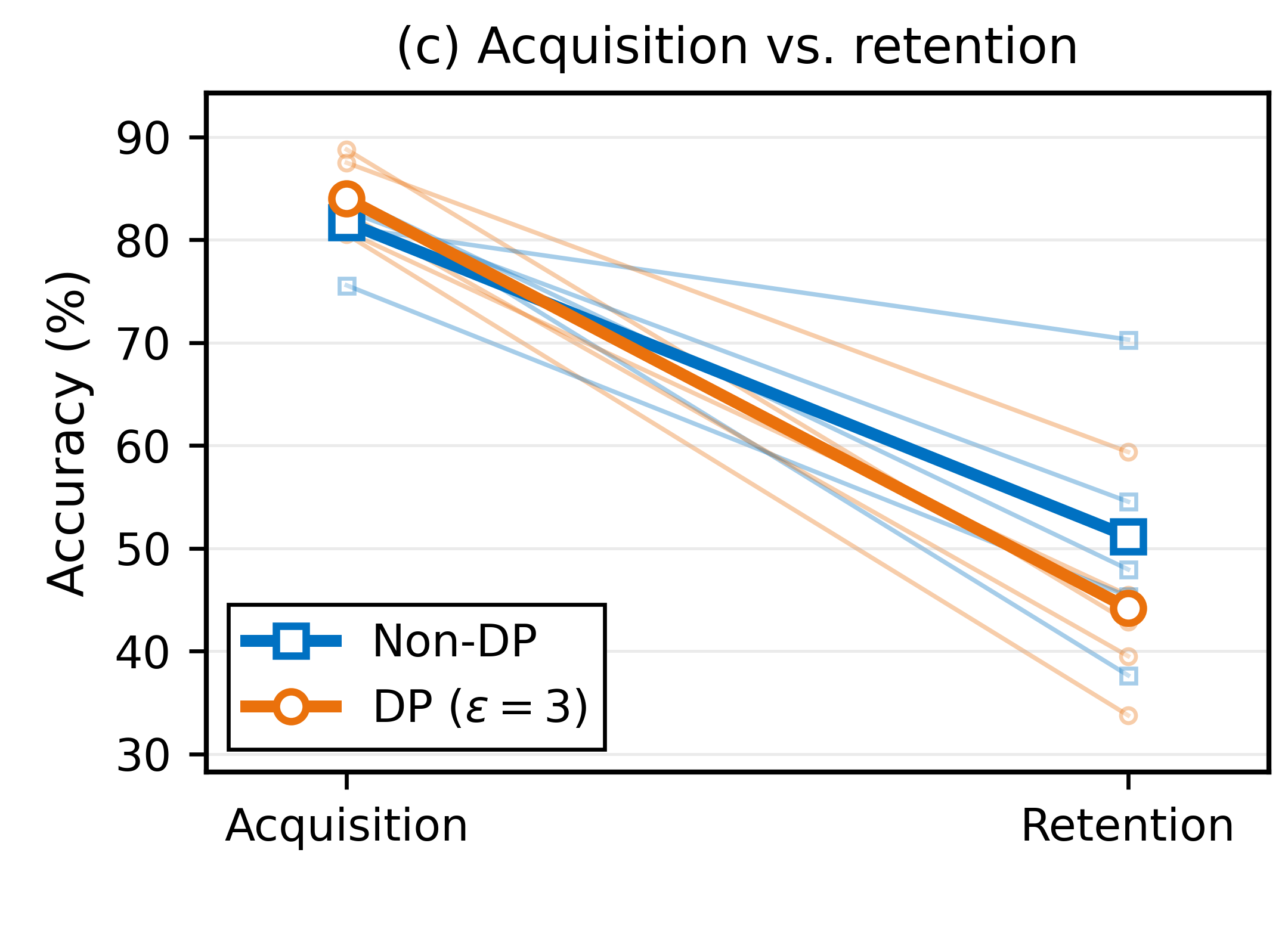}
        \label{fig:challenge1_c}
    \end{subfigure}
    \vspace{-3mm}
    \caption{DP exacerbates forgetting during continual adaptation.
(a) Average forgetting across tasks. (b) Early-task forgetting over subsequent adaptations.
(c) Acquisition vs. retention after 15 subsequent adaptations. Light lines denote individual tasks; bold lines denote their mean.}
    \label{fig:challenge_forgetting}
\end{figure*}

\noindent
\textbf{Observation 1.} 
\textit{Differential privacy substantially exacerbates forgetting in low-rank continual adaptation, with the degradation arising mainly during subsequent adaptation rather than initial task acquisition.} 
Let $a_{t,k}$ denote the accuracy on task $k$ after learning task $t$.
We measure forgetting as
$F_k^t=\max_{\tau\in\{k,\ldots,t-1\}}a_{\tau,k}-a_{t,k}$ and
$F^t=\frac{1}{t-1}\sum_{k=1}^{t-1}F_k^t$. As shown in Fig.~\ref{fig:challenge_forgetting}(a), DP ($\epsilon=3$) incurs substantially larger forgetting than Non-DP throughout the task sequence. To examine how this degradation evolves, we fix the first $K=5$ tasks and track their forgetting after $u$ subsequent adaptations,
$\bar{F}^{u}=\frac{1}{K}\sum_{k=1}^{K}F_k^{k+u}$.
Fig.~\ref{fig:challenge_forgetting}(b) shows that forgetting increases under both settings, while remaining consistently higher under DP.
Fig.~\ref{fig:challenge_forgetting}(c) further compares acquisition accuracy $a_{k,k}$ with retention accuracy $a_{k+15,k}$. Although acquisition is comparable between DP and Non-DP, retention falls to 44.20\% under DP versus 51.17\% under Non-DP after 15 subsequent adaptations, indicating that the additional forgetting emerges mainly during later adaptation.

\noindent
\textbf{Observation 2.}
\textit{Jointly privatizing the two LoRA factors introduces a non-negligible bilinear noise interaction during sequential adaptation.}
Consider one DP-SGD update within a task. Let $\boldsymbol A$ and $\boldsymbol B$ denote the LoRA factors after the corresponding noise-free clipped-gradient update, and let $\boldsymbol\xi_A$ and $\boldsymbol\xi_B$ denote the parameter perturbations induced by the Gaussian gradient noise in this update. The resulting factors can be written as $ \widetilde{\boldsymbol A} = \boldsymbol A+\boldsymbol\xi_A, \widetilde{\boldsymbol B} = \boldsymbol B+\boldsymbol\xi_B.$ Accordingly, the privacy-induced distortion in the effective LoRA update is 
\begin{equation} \widetilde{\boldsymbol B}\widetilde{\boldsymbol A} - \boldsymbol B\boldsymbol A = \boldsymbol B\boldsymbol\xi_A + \boldsymbol\xi_B\boldsymbol A + \boldsymbol\xi_B\boldsymbol\xi_A .  
\end{equation} 
The first two terms describe the first-order propagation of the factor-wise perturbations, whereas $\boldsymbol\xi_B\boldsymbol\xi_A$ captures their second-order interaction induced by the bilinear LoRA parameterization.

\begin{wrapfigure}{r}{0.35\textwidth}
    \centering
    \vspace{-0pt}
    \includegraphics[width=\linewidth]{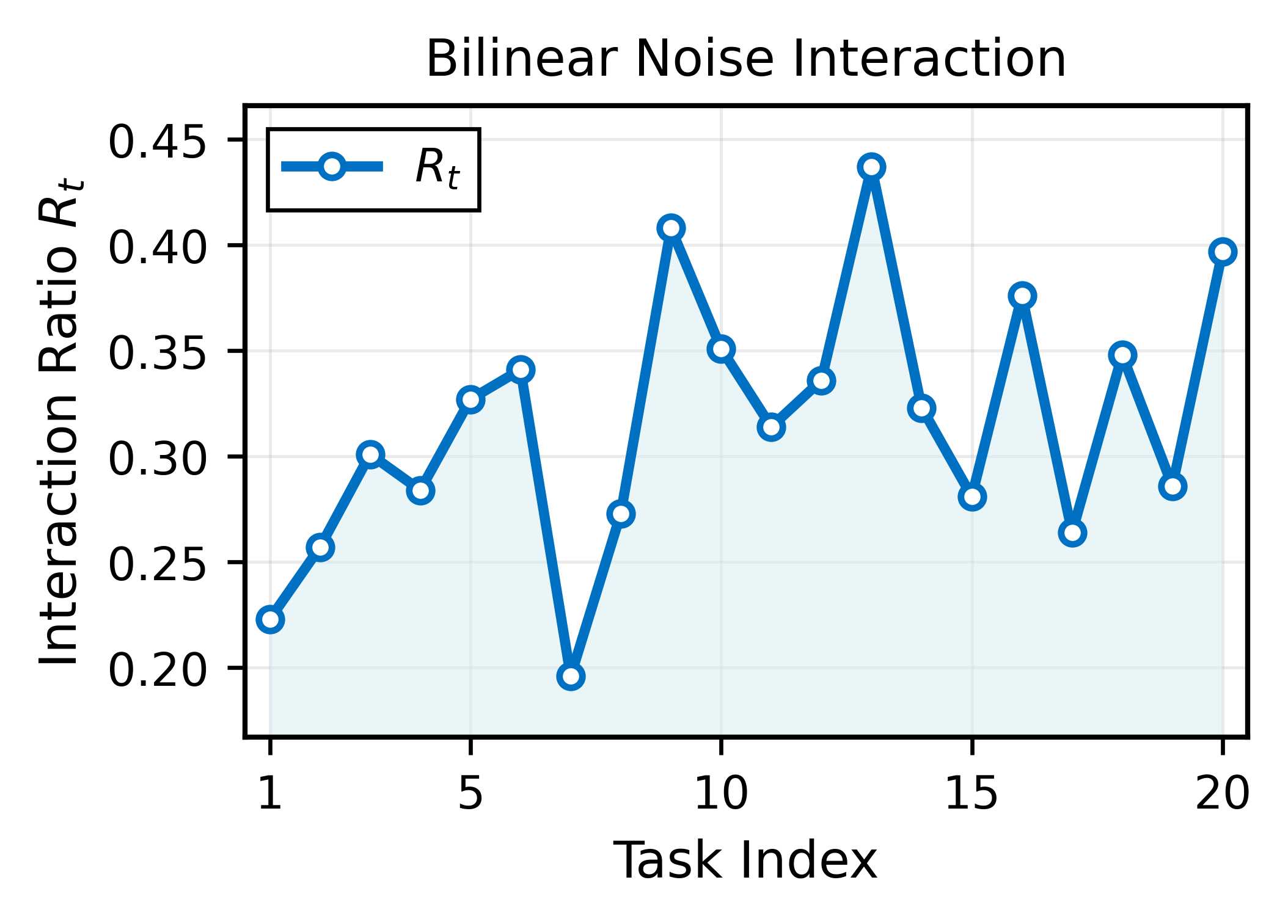}
    \vspace{-7mm}
    \caption{Bilinear interaction ratio.}
    \label{fig:challenge_bilinear}
    \vspace{-25pt}
\end{wrapfigure} 
We quantify the relative magnitude of the bilinear interaction over the continual sequence under $\sigma=0.8$. We aggregate the perturbation energies over DP-SGD steps and LoRA modules:
{\small
\begin{equation}
R_t=
\frac{
\sqrt{\sum_{\ell,q}
\|\boldsymbol\xi_{B,t,q}^{(\ell)}
\boldsymbol\xi_{A,t,q}^{(\ell)}\|_F^2}
}{
\sqrt{\sum_{\ell,q}
\|\boldsymbol B_{t,q}^{(\ell)}
\boldsymbol\xi_{A,t,q}^{(\ell)}
+
\boldsymbol\xi_{B,t,q}^{(\ell)}
\boldsymbol A_{t,q}^{(\ell)}\|_F^2}
},
\end{equation}}where $q$ indexes mini-batch DP-SGD updates within task $t$ and $\ell$ indexes LoRA modules.
The bilinear interaction remains consistently non-negligible throughout the 20-task sequence, despite substantial task-wise variation (Fig.~\ref{fig:challenge_bilinear}).

\begin{figure*}[h]
    \centering
    \begin{subfigure}[t]{0.32\textwidth}
        \centering \includegraphics[width=\linewidth]{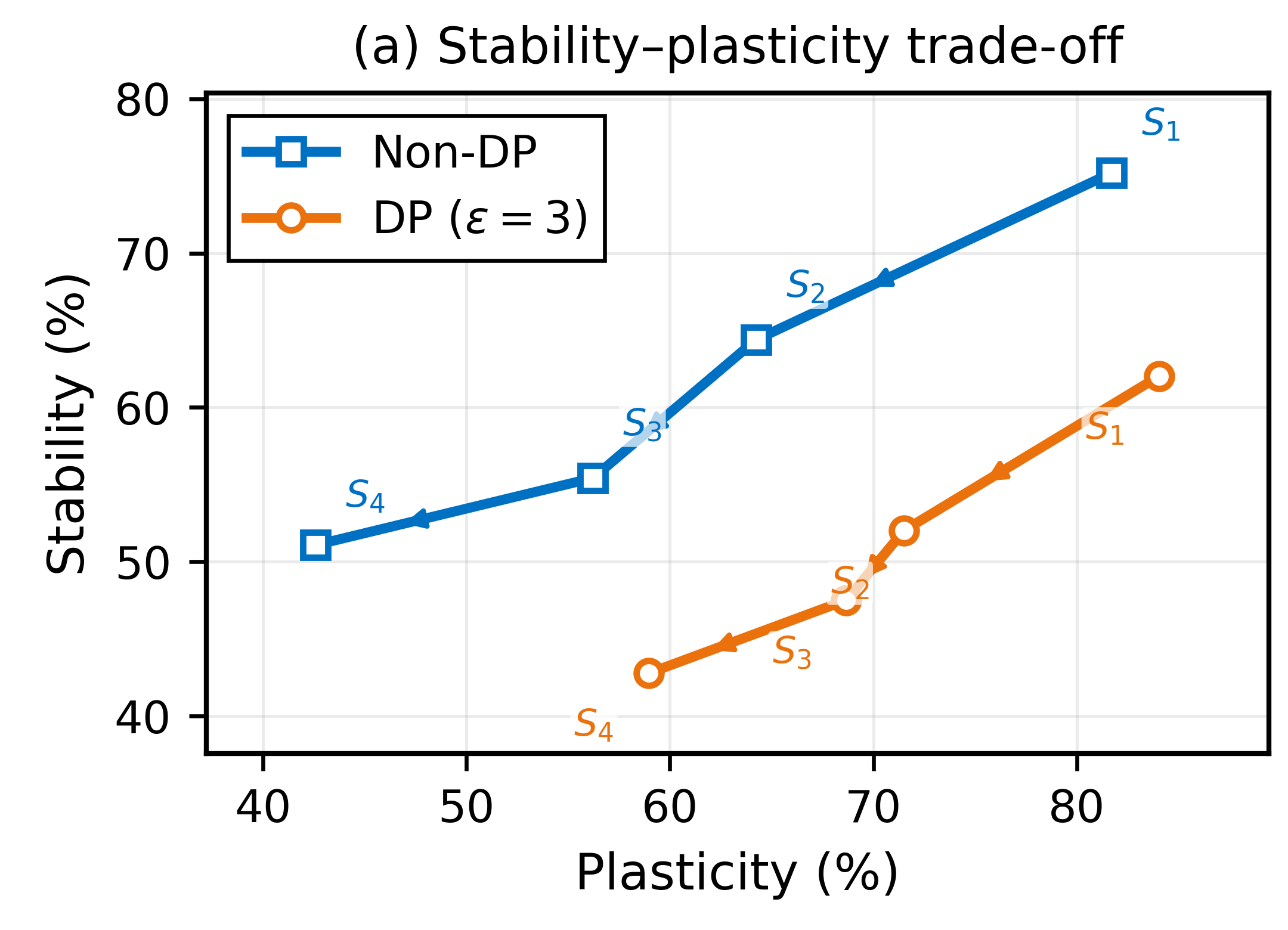}
    \end{subfigure}
    \hfill
    \begin{subfigure}[t]{0.32\textwidth}
        \centering
\includegraphics[width=\linewidth]{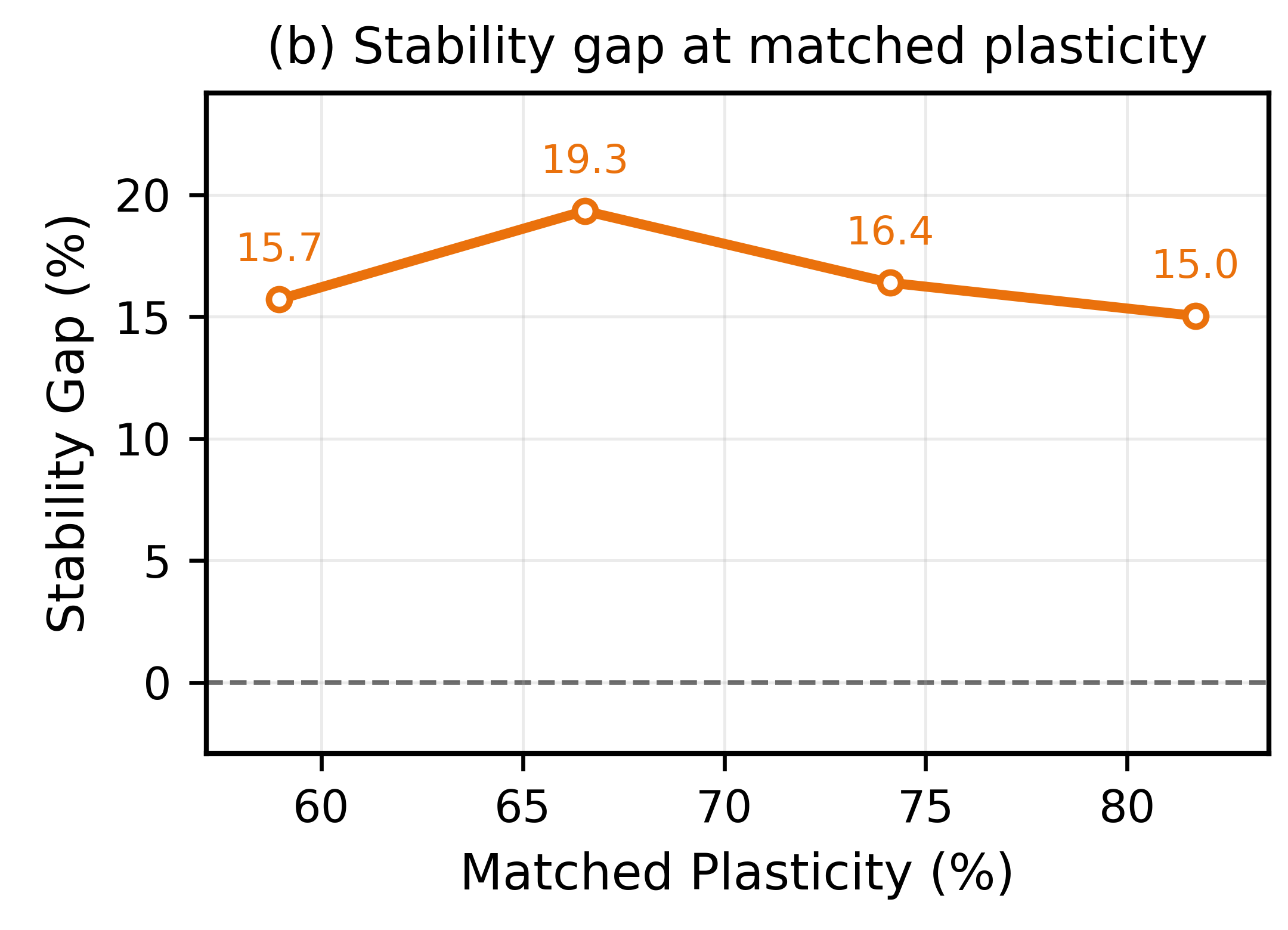}
    \end{subfigure}
    \hfill
    \begin{subfigure}[t]{0.32\textwidth}
        \centering
\includegraphics[width=\linewidth]{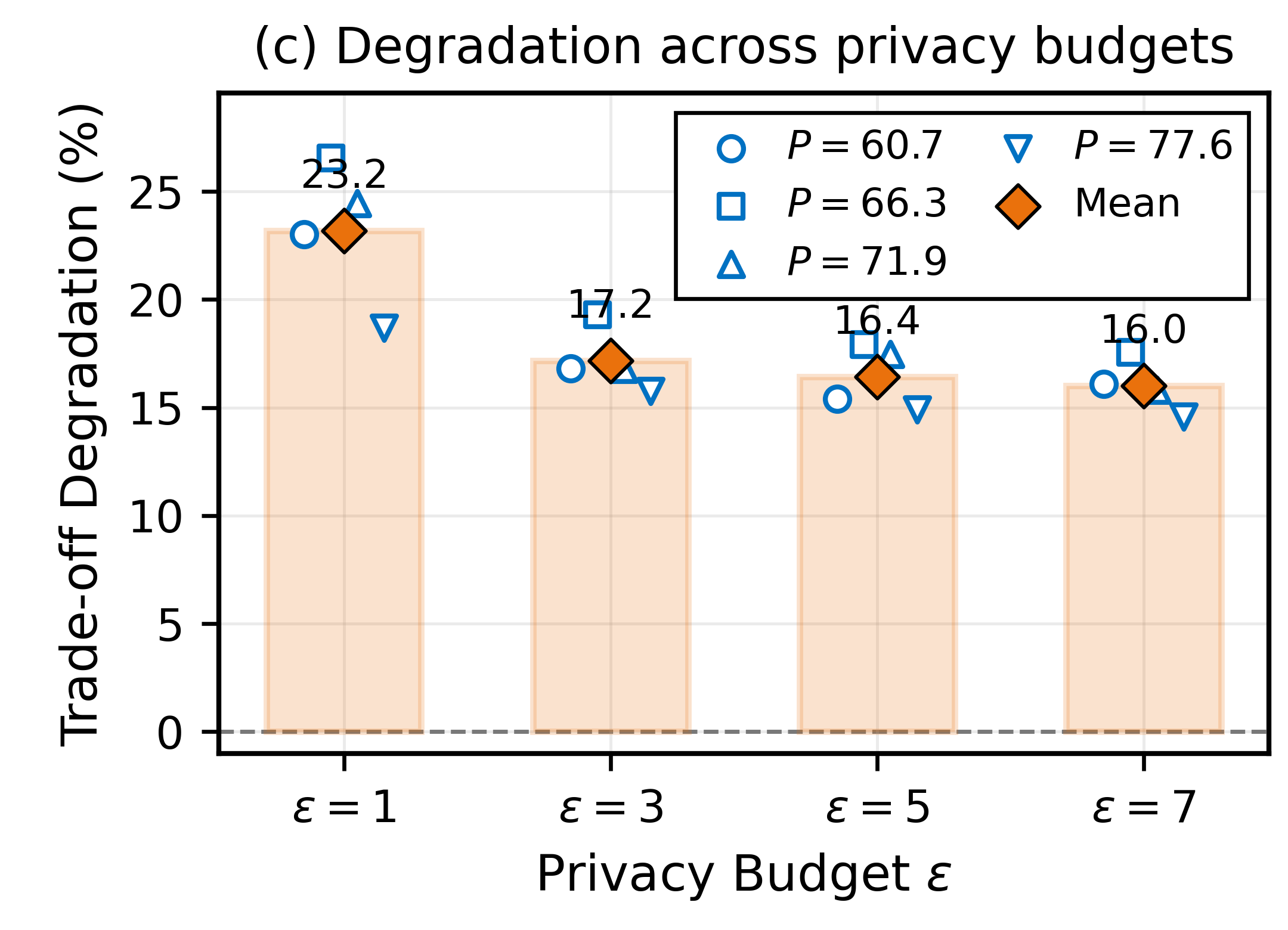}
    \end{subfigure}
    \vspace{-2mm}
    \caption{
Stability--plasticity trade-off under differential privacy.
(a) Stage-wise stability--plasticity trajectories for Non-DP and DP ($\epsilon=3$). (b) Stability gap between Non-DP and DP at matched plasticity. (c) Trade-off degradation across privacy budgets, where markers denote stability gaps at individual matched plasticity levels $P$ and bars denote their mean.
}
    \label{fig:challenge_3}
\end{figure*}

\noindent
\textbf{Observation 3.}
\textit{Differential privacy degrades the stability--plasticity trade-off, making it harder to retain previous knowledge while effectively adapting to new tasks.}
We characterize plasticity $P$ by new-task accuracy and stability by old-task accuracy. We divide the 20-task sequence into four consecutive stages of five tasks
each and average the new-task and old-task accuracies within each stage
to obtain the stage-wise plasticity--stability operating points.
As shown in Fig.~\ref{fig:challenge_3}(a), both DP and Non-DP shift toward lower plasticity and stability as adaptation progresses, while DP consistently operates in a lower-stability region. To isolate the stability cost from differences in new-task adaptation,
Fig.~\ref{fig:challenge_3}(b) compares the two settings at matched plasticity levels.
The stability gap remains positive across the matched range, showing that comparable plasticity under DP is associated with systematically lower stability.
Fig.~\ref{fig:challenge_3}(c) further shows that this trade-off degradation persists across privacy budgets and becomes less severe as the privacy constraint is relaxed.

\vspace{-3mm}
\section{Proposed Method}
\label{sec:method}

\begin{figure}[h]
    \centering
    \includegraphics[width=0.98\textwidth]{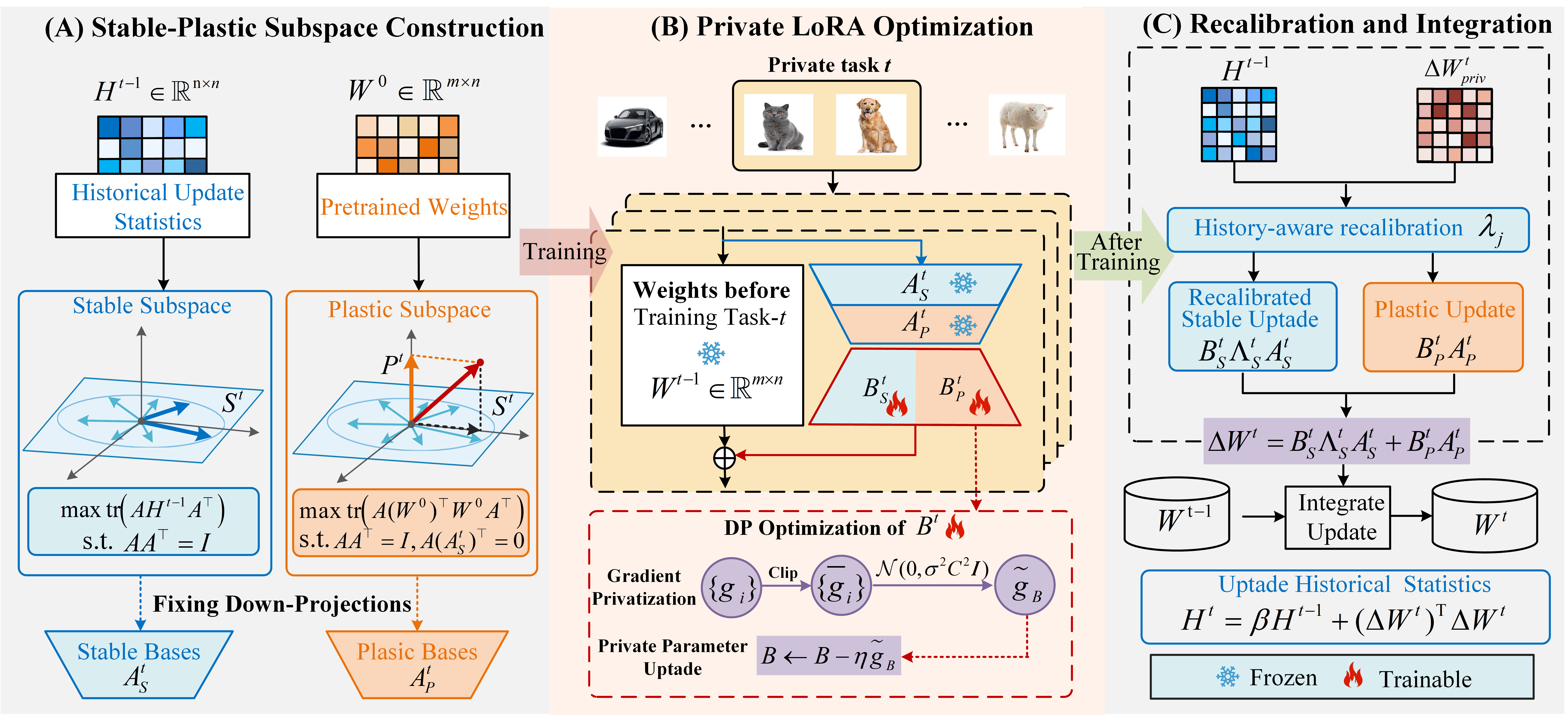}
\caption{
Overview of \textsf{DP-SPA}.
(A) Construct the stable subspace $\mathcal S^t$ from privatized
update history $\boldsymbol H^{t-1}$ and the plastic subspace
$\mathcal P^t$ from pretrained weights $\boldsymbol W^0$
within $(\mathcal S^t)^\perp$.
(B) Within task $t$, fix $\boldsymbol A_S^t,\boldsymbol A_P^t$
and optimize only $\boldsymbol B_S^t,\boldsymbol B_P^t$ with DP-SGD.
(C) Recalibrate stable-branch updates using historical and current
update energies, merge $\Delta\boldsymbol W^t$ into the model,
and update $\boldsymbol H^t$ for the next task.}
    \label{fig:subspace_evolution}
    \vspace{-2mm}
\end{figure}

\noindent
\textbf{Overview.} To address the issues in Section~\ref{subsec:challenges}, we propose \textsf{DP-SPA}, which constructs stable and plastic
LoRA bases before accessing current-task private data and restricts direct access to these data to DP-SGD. At task $t$, the stable subspace $\mathcal S^t$ captures dominant historical adaptation directions from
$\boldsymbol H^{t-1}$, a statistic of previously privatized model updates.
The plastic subspace $\mathcal P^t$ is constructed from
the pretrained weight $\boldsymbol W^0$ within the orthogonal complement of $\mathcal S^t$, providing complementary directions for new-task adaptation (Fig.~\ref{fig:subspace_evolution}(A)).
The down-projection matrices $\boldsymbol A_S^t$ and
$\boldsymbol A_P^t$ remain fixed, while only the corresponding up-projection matrices $\boldsymbol B_S^t$ and $\boldsymbol B_P^t$ are trained using DP-SGD, eliminating the bilinear noise interaction
(Fig.~\ref{fig:subspace_evolution}(B)).
After private optimization, we recalibrate the stable-branch update using historical update energy and current update energy computed from the DP-SGD outputs.
The resulting LoRA update $\Delta\boldsymbol W^t$ is merged into the continual model and incorporated into
$\boldsymbol H^t$ to guide stable-subspace construction
for the next task (Fig.~\ref{fig:subspace_evolution}(C)).

\subsection{Privacy-Constrained Stable-Plastic Subspace Construction}
\label{sec:sp_subspace}

At task $t$, the LoRA adaptation is parameterized by
$\boldsymbol B^t\boldsymbol A^t$.
$\boldsymbol A^t\in\mathbb R^{r\times n}$
determines the low-dimensional input-side subspace through which adaptation is expressed, while
$\boldsymbol B^t\in\mathbb R^{m\times r}$
maps this representation to the output space~\cite{zhu2024asymmetry,tian2024hydralora}.
We therefore use $\boldsymbol A^t$ to determine the low-rank adaptation subspace. Under continual adaptation, the limited rank budget must preserve historically occupied directions while retaining capacity outside them for subsequent tasks.
We therefore decompose the row space of
$\boldsymbol A^t$
into orthogonal stable and plastic components:
\vspace{-1mm}
\begin{equation}
    \boldsymbol A^t
    =
    \begin{bmatrix}
        \boldsymbol A_S^t\\
        \boldsymbol A_P^t
    \end{bmatrix},
    \qquad
    r_S+r_P=r,
    \label{eq:sp_decomposition}
\end{equation}
where the rows of
$\boldsymbol A_S^t\in\mathbb{R}^{r_S\times n}$ and $\boldsymbol A_P^t\in\mathbb{R}^{r_P\times n}$
span the stable subspace $\mathcal{S}^t$ and plastic subspace $\mathcal{P}^t$. The ranks $r_S$ and $r_P$ determine the capacity allocated to the two components. Both bases are constructed before accessing the current private dataset
and satisfy
$\boldsymbol A_S^t(\boldsymbol A_S^t)^\top=\boldsymbol I$,
$\boldsymbol A_P^t(\boldsymbol A_P^t)^\top=\boldsymbol I$, and
$\boldsymbol A_P^t(\boldsymbol A_S^t)^\top=\boldsymbol 0$.
Accordingly, the LoRA up-projection is partitioned as $\boldsymbol{B}^t=[\boldsymbol{B}_S^t,\boldsymbol{B}_P^t]$.

\paragraph{Historical stable subspace.}
We construct the stable subspace from the updates that have already been privatized and merged into the continual model. To summarize their directional usage, we maintain the exponentially weighted second-order statistic 
$
    \boldsymbol H^{t-1}
    =
    \sum\nolimits_{i=1}^{t-1}
    \beta^{\,t-1-i}
    (\Delta\boldsymbol W^i)^\top
    \Delta\boldsymbol W^i,
    \label{eq:historical_statistic}
$ 
where $\beta\in(0,1]$ controls the decay of earlier task updates.
For any unit direction $\boldsymbol a\in\mathbb R^n$, $ 
    \boldsymbol a^\top
    \boldsymbol H^{t-1}
    \boldsymbol a
    =
    \sum_{i=1}^{t-1}
    \beta^{\,t-1-i}
    \left\|
        \Delta\boldsymbol W^i\boldsymbol a
    \right\|_2^2.
$ 
Hence, $\boldsymbol H^{t-1}$ summarizes the accumulated update energy along different right-side directions. We define the stable subspace as the $r_S$-dimensional subspace that preserves the largest amount of this accumulated historical update energy:
\vspace{-0mm}
\begin{equation}
    \boldsymbol A_S^t
    = \arg\max_{\boldsymbol A}
    \operatorname{tr}\!\left(
    \boldsymbol A\boldsymbol H^{t-1}\boldsymbol A^\top
    \right)
    \quad \text{s.t.}\quad
    \boldsymbol A\boldsymbol A^\top=\boldsymbol I.
    \label{eq:stable_objective}
\end{equation}
Here, $\boldsymbol A\in\mathbb R^{r_S\times n}$,
and $\arg\max$ denotes a selected maximizer. Let
$\boldsymbol H^{t-1}
=\boldsymbol V_H^t\boldsymbol\Lambda_H^t
(\boldsymbol V_H^t)^\top$
be its eigendecomposition, where $\boldsymbol V_H^t$ contains orthonormal eigenvectors and $\boldsymbol\Lambda_H^t=\operatorname{diag}(h_1^t,\ldots,h_n^t)$ with $h_1^t\ge\cdots\ge h_n^t\ge0$, the solution is
$\boldsymbol A_S^t=(\boldsymbol V_{H,1:r_S}^t)^\top$
by the Rayleigh--Ritz principle.
Thus, $\mathcal S^t$ captures the dominant directions of historical adaptation.

\paragraph{Residual plastic subspace.}
To reserve adaptation capacity outside the historical stable
subspace, we allocate the remaining $r_P$ directions to its
orthogonal complement.
Since the basis is constructed before
accessing the current private dataset, we use the pretrained weight $\boldsymbol W^0$ as a structural prior. Specifically, we select the subspace with maximal pretrained response energy:
\vspace{-1mm}
\begin{equation}
    \boldsymbol A_P^t
= \arg\max_{\boldsymbol A}
\operatorname{tr}\!\left(
\boldsymbol A(\boldsymbol W^0)^\top
\boldsymbol W^0\boldsymbol A^\top
\right)
\quad \mathrm{s.t.}\quad
\boldsymbol A\boldsymbol A^\top=\boldsymbol I,\;
\boldsymbol A(\boldsymbol A_S^t)^\top=\boldsymbol 0.
    \label{eq:plastic_objective}
\end{equation}Here, $\boldsymbol A\in\mathbb R^{r_P\times n}$.
Let $\boldsymbol U_\perp^t\in\mathbb R^{n\times(n-r_S)}$ be an orthonormal basis of $(\mathcal S^t)^\perp$. Setting $\boldsymbol A=\boldsymbol Z^\top(\boldsymbol U_\perp^t)^\top$, where $\boldsymbol Z^\top\boldsymbol Z=\boldsymbol I_{r_P}$, reduces Eq.~\ref{eq:plastic_objective} to maximizing $\operatorname{tr}(\boldsymbol Z^\top\boldsymbol C_P^t\boldsymbol Z)$, where
$ 
    \boldsymbol C_P^t
    =
    (\boldsymbol U_\perp^t)^\top
    (\boldsymbol W^0)^\top\boldsymbol W^0
    \boldsymbol U_\perp^t.
\label{eq:restricted_pretrained_gram}
$
Let $\mu_1^t\ge\cdots\ge\mu_{n-r_S}^t\ge0$ denote the eigenvalues of $\boldsymbol C_P^t$.
For $r_P\leq n{-}r_S$, let $\boldsymbol Z_{P,1:r_P}^t$
contain orthonormal eigenvectors associated with the
$r_P$ largest eigenvalues $\mu_1^t,\ldots,\mu_{r_P}^t$ of $\boldsymbol C_P^t$. Then
\begin{equation}
    \boldsymbol A_P^t
    =
    (\boldsymbol Z_{P,1:r_P}^t)^\top
    (\boldsymbol U_\perp^t)^\top.
    \label{eq:plastic_basis}
\end{equation}
Thus,
$\boldsymbol A_P^t(\boldsymbol A_P^t)^\top
=\boldsymbol I_{r_P}$ and
$\boldsymbol A_P^t(\boldsymbol A_S^t)^\top=\boldsymbol 0$,
including when $\boldsymbol C_P^t$ is rank deficient.
The resulting plastic subspace maximizes pretrained response
energy within $(\mathcal S^t)^\perp$.


\noindent\textbf{Initialization.}
For task $t=1$, we initialize the fixed basis using the dominant right-singular directions of the pretrained weight:
$\boldsymbol W^0=\boldsymbol U_0\boldsymbol\Sigma_0\boldsymbol V_0^\top$,
$\boldsymbol A^1=\boldsymbol V_{0,1:r}^\top$.
After task 1, $\Delta\boldsymbol W^1$ initializes $\boldsymbol H^1$,
enabling the stable--plastic construction for subsequent tasks.

\subsection{Private LoRA Optimization}
\label{sec:private_optimization}

Once the stable--plastic basis is constructed, we keep
$\boldsymbol A^t$ fixed throughout task $t$ and privately optimize only $\boldsymbol B^t$ within each LoRA module. Fixing $\boldsymbol A^t$ preserves the adaptation geometry
constructed from historical and pretrained information
and eliminates the bilinear noise interaction identified
in \textbf{Observation~2}. We apply the DP-SGD mechanism defined in
Section~\ref{subsec:problem_formulation} to $\boldsymbol B^t$:
$ 
    \boldsymbol B^t
    \leftarrow
    \boldsymbol B^t
    -
    \eta\,\widetilde{\boldsymbol g}_{B}^{\,t},
    \label{eq:private_B_update}
$
where $\eta>0$ is the learning rate and $\widetilde{\boldsymbol g}_{B}^{\,t}$ denotes the DP-SGD gradient with respect to $\boldsymbol B^t$. Following the single-step perturbation analysis in~\textbf{Observation~2}, let
$\boldsymbol\xi_B^t$ denote the Gaussian perturbation applied to
$\boldsymbol B^t$.
Because $\boldsymbol A^t$ is fixed, the induced distortion in the
effective LoRA update reduces to
$ 
    (\boldsymbol B^t+\boldsymbol\xi_B^t)\boldsymbol A^t
    -
    \boldsymbol B^t\boldsymbol A^t
    =
    \boldsymbol\xi_B^t\boldsymbol A^t.
    \label{eq:single_factor_perturbation}
$ 
Compared with joint privatization, the perturbation terms associated with the down-projection,
$\boldsymbol B^t\boldsymbol\xi_A^t$ and
$\boldsymbol\xi_B^t\boldsymbol\xi_A^t$,
are eliminated. Hence, private perturbation propagates only through the fixed adaptation basis, without basis noise or bilinear noise interaction between
the two LoRA factors.

\subsection{History-Aware Recalibration and Continual Integration}
\label{sec:history_integration}

After private optimization, the resulting LoRA update is decomposed as
$\Delta\boldsymbol W_{\mathrm{priv}}^t
=\boldsymbol B_S^t\boldsymbol A_S^t
+\boldsymbol B_P^t\boldsymbol A_P^t$.
Stable directions differ in their historical update energy
and current adaptation strength.
We therefore recalibrate their contributions individually
before model integration.
For the $j$-th stable direction, we define its historical update energy as
$h_j^t=\boldsymbol a_{S,j}^t\boldsymbol H^{t-1}
(\boldsymbol a_{S,j}^t)^\top$. Here, $\boldsymbol a_{S,j}^t$ denotes the $j$-th row of $\boldsymbol A_S^t$. Since $\boldsymbol a_{S,j}^t$ is an eigenvector of $\boldsymbol H^{t-1}$, $h_j^t$ equals its associated eigenvalue. We measure the current adaptation strength by the energy
of the corresponding rank-one update,
$c_j^t=\|\boldsymbol b_{S,j}^t\boldsymbol a_{S,j}^t\|_F^2
=\|\boldsymbol b_{S,j}^t\|_2^2$,
where $\boldsymbol b_{S,j}^t$ is the $j$-th column of
$\boldsymbol B_S^t$ and
$\|\boldsymbol a_{S,j}^t\|_2=1$.

We determine a recalibration coefficient by balancing fidelity to the privately learned update against modification along historically active directions.
Specifically, for each stable direction we solve
$c_j^t(1-\lambda)^2+(h_j^t+\tau)\lambda^2$
over $\lambda\in[0,1]$, where $\tau>0$ is a small
regularization constant.
The first term penalizes deviation from the privately learned rank-one update, whereas the second penalizes retaining a large current update along a historically active direction.
The unique minimizer is
\begin{equation}
    \lambda_j^t
    ={c_j^t}/({h_j^t+c_j^t+\tau}),
    \qquad
    0\leq\lambda_j^t<1.
    \label{eq:recalibration_coefficient}
\end{equation}When historical update energy dominates, the corresponding current stable-direction update is attenuated more strongly. When the current adaptation strength is comparatively larger, a greater fraction of the privately learned update is retained.

Let $\boldsymbol\Lambda_S^t
=\operatorname{diag}(\lambda_1^t,\ldots,\lambda_{r_S}^t)$,
the final update is then 
\begin{equation}
    \Delta\boldsymbol W^t =\boldsymbol B_S^t\boldsymbol\Lambda_S^t\boldsymbol A_S^t+\boldsymbol B_P^t\boldsymbol A_P^t.
    \label{eq:recalibrated_update}
\end{equation}
The plastic branch is retained without recalibration.

For $t=1$, we use $\Delta\boldsymbol W^1=\boldsymbol B^1\boldsymbol A^1$ without recalibration.
At every task, we merge the update via
$\boldsymbol W^t=\boldsymbol W^{t-1}+\Delta\boldsymbol W^t$
and update the historical statistic recursively:
$
\boldsymbol H^t
=
\beta \boldsymbol H^{t-1}
+
(\Delta\boldsymbol W^t)^\top
\Delta\boldsymbol W^t.
\label{eq:H_t}
$
Hence, $\boldsymbol H^t$ summarizes the accumulated energy of the updates merged into the continual model and guides stable-subspace construction for task $t+1$.

\section{Theoretical Analysis}
\label{sec:theoretical_analysis}
We establish the privacy guarantee, characterize
the stable--plastic geometry, and analyze recalibration and its conditional effect on old-task loss. Complete proofs are deferred to Appendix~\ref{app:Theoretical}.

\begin{theorem}[Privacy Guarantee]
\label{thm:dp_main}
Assume public pretrained weights and a fixed public task partition, with each protected record belonging to exactly one task. Under add/remove adjacency, suppose each task mechanism is $(\epsilon_t,\delta_t)$-DP conditional on any preceding private transcript, and all remaining operations use only public information and privatized outputs. Then the full transcript of released models and statistics is
$(\max_t\epsilon_t,\max_t\delta_t)$-DP.
In particular, assigning the same
$(\epsilon,\delta)$ budget to every task gives an
$(\epsilon,\delta)$ guarantee for the entire sequence.
\end{theorem}
Theorem~\ref{thm:dp_main} allows later tasks to depend
adaptively on earlier private outputs.
Since basis construction, recalibration, merging, and history
updates use only public information and previously privatized
outputs, they constitute post-processing and incur no additional
privacy cost.

For any update $\boldsymbol U\in\mathbb R^{m\times n}$, define the historical overlap energy
$
\mathcal{I}_t(\boldsymbol U)
=
\operatorname{tr}\!\left(
\boldsymbol U\boldsymbol H^{t-1}\boldsymbol U^\top
\right)
=
\sum_{i<t}\beta^{t-1-i}
\|\boldsymbol U(\Delta\boldsymbol W^i)^\top\|_F^2,
$
which measures its energy along historical update directions.

\begin{theorem}[Complementary Stable--Plastic Geometry]
\label{thm:geometry_main}
For $t\ge2$,
$\boldsymbol A^t(\boldsymbol A^t)^\top=\boldsymbol I_r$,
and the following statements hold.
(i) Historical energy.
Over all $\boldsymbol U\in\mathbb R^{r_S\times n}$ satisfying
$\boldsymbol U\boldsymbol U^\top=\boldsymbol I_{r_S}$, the stable basis satisfies
$
\max_{\boldsymbol U}
\operatorname{tr}\!\left(
\boldsymbol U\boldsymbol H^{t-1}\boldsymbol U^\top
\right)
=
\operatorname{tr}\!\left(
\boldsymbol A_S^t\boldsymbol H^{t-1}(\boldsymbol A_S^t)^\top
\right)
=
\sum_{j=1}^{r_S}h_j^t.
$
Every plastic update obeys
$
\mathcal I_t(\boldsymbol B_P^t\boldsymbol A_P^t)
\le
h_{r_S+1}^t\|\boldsymbol B_P^t\|_F^2.
$
(ii) Pretrained response.
Over all $\boldsymbol U\in\mathbb R^{r_P\times n}$ satisfying
$\boldsymbol U\boldsymbol U^\top=\boldsymbol I_{r_P}$ and
$\boldsymbol U(\boldsymbol A_S^t)^\top=\boldsymbol 0$,
the plastic basis satisfies
$
\max_{\boldsymbol U}
\|\boldsymbol W^0\boldsymbol U^\top\|_F^2
=
\|\boldsymbol W^0(\boldsymbol A_P^t)^\top\|_F^2
=
\sum_{j=1}^{r_P}\mu_j^t.
$
(iii) Additional adaptation capacity.
For every target update
$\boldsymbol U\in\mathbb R^{m\times n}$,
$
\min_{\boldsymbol X_S\in\mathbb R^{m\times r_S}}
\|\boldsymbol U-\boldsymbol X_S\boldsymbol A_S^t\|_F^2
-
\min_{\boldsymbol X\in\mathbb R^{m\times r}}
\|\boldsymbol U-\boldsymbol X\boldsymbol A^t\|_F^2
=
\|\boldsymbol U(\boldsymbol A_P^t)^\top\|_F^2.
$
Moreover,
$\|\boldsymbol X\boldsymbol A^t\|_F=\|\boldsymbol X\|_F$
for every
$\boldsymbol X\in\mathbb R^{m\times r}$.
\end{theorem}

\begin{theorem}[Optimal Recalibration and Historical Overlap Reduction]
\label{thm:recalibration_main}
Condition on the realized private coefficient matrix $\boldsymbol B^t$ and history $\boldsymbol H^{t-1}$. For stable columns with $c_j^t=0$, impose $\boldsymbol x_j=\boldsymbol 0$, where $\boldsymbol x_j$ denotes the $j$-th column of $\boldsymbol X_S$. Then $\widehat{\boldsymbol B}^t=[\boldsymbol B_S^t\boldsymbol\Lambda_S^t,\boldsymbol B_P^t]$ is the unique minimizer of $\min_{\boldsymbol X=[\boldsymbol X_S,\boldsymbol X_P]}\bigl\{\frac12\|\boldsymbol X-\boldsymbol B^t\|_F^2+\frac12\sum_{j:c_j^t>0}\frac{h_j^t+\tau}{c_j^t}\|\boldsymbol x_j\|_2^2\bigr\}$. Furthermore, $\Delta\boldsymbol W^t(\boldsymbol A_P^t)^\top=\boldsymbol B_P^t$, $\|\Delta\boldsymbol W_{\mathrm{priv}}^t-\Delta\boldsymbol W^t\|_F^2=\sum_{j=1}^{r_S}c_j^t(1-\lambda_j^t)^2$, $\|\Delta\boldsymbol W_{\mathrm{priv}}^t\|_F^2-\|\Delta\boldsymbol W^t\|_F^2=\sum_{j=1}^{r_S}c_j^t(1-(\lambda_j^t)^2)$, and $\mathcal{I}_t(\Delta\boldsymbol W_{\mathrm{priv}}^t)-\mathcal{I}_t(\Delta\boldsymbol W^t)=\sum_{j=1}^{r_S}h_j^tc_j^t(1-(\lambda_j^t)^2)\ge0$. 
\end{theorem}
Recalibration balances fidelity to the private update
against an energy-normalized historical penalty,
preserving plastic coefficients without increasing
historical overlap.

\begin{theorem}[Conditional Forgetting Certificate]
\label{thm:forgetting_main}
Let $F_{\mathrm{old}}$ be an old-task population loss and let $\boldsymbol W^{t-1}$ be the pre-merge weight. Suppose $\|\nabla F_{\mathrm{old}}(\boldsymbol W^{t-1})\|_F\le\nu$. Assume that $F_{\mathrm{old}}$ is twice continuously differentiable along the relevant segments and that, for $\boldsymbol U\in\{\Delta\boldsymbol W_{\mathrm{priv}}^t,\Delta\boldsymbol W^t\}$, $s\in[0,1]$, and every perturbation $\boldsymbol V\in\mathbb R^{m\times n}$, $\nabla^2F_{\mathrm{old}}(\boldsymbol W^{t-1}+s\boldsymbol U)[\boldsymbol V,\boldsymbol V]\le\kappa\operatorname{tr}(\boldsymbol V\boldsymbol H^{t-1}\boldsymbol V^\top)+\rho\|\boldsymbol V\|_F^2$, where $\nu,\kappa,\rho\ge0$. Define $\mathcal{U}(\boldsymbol U)=\nu\|\boldsymbol U\|_F+\frac{\kappa}{2}\mathcal{I}_t(\boldsymbol U)+\frac{\rho}{2}\|\boldsymbol U\|_F^2$. Then $F_{\mathrm{old}}(\boldsymbol W^{t-1}+\boldsymbol U)-F_{\mathrm{old}}(\boldsymbol W^{t-1})\le\mathcal{U}(\boldsymbol U)$ for $\boldsymbol U\in\{\Delta\boldsymbol W_{\mathrm{priv}}^t,\Delta\boldsymbol W^t\}$, and $\mathcal{U}(\Delta\boldsymbol W_{\mathrm{priv}}^t)-\mathcal{U}(\Delta\boldsymbol W^t)=\nu\bigl(\|\Delta\boldsymbol W_{\mathrm{priv}}^t\|_F-\|\Delta\boldsymbol W^t\|_F\bigr)+\frac12\sum_{j=1}^{r_S}(\kappa h_j^t+\rho)c_j^t(1-(\lambda_j^t)^2)\ge0$.
\end{theorem}
Under the stated assumptions, recalibration does not increase the upper bound on old-task loss increase, but need not reduce the actual loss. The same argument extends to multiple adapted layers under a joint curvature bound accounting for cross-layer interactions.

\section{Experiments}
\subsection{Experimental Setups}
\label{sec:Experimental Setups}
\textbf{Evaluation Benchmarks and Metrics.} We evaluate
DP-SPA on 5 benchmarks: ImageNet-R~\citep{hendrycks2021many}, ImageNet-A~\citep{hendrycks2021natural}, CIFAR-100~\citep{krizhevsky2009learning}, CUB~\citep{wah2011caltech} and DomainNet~\citep{peng2019moment}. ImageNet-R contains 30,000 images from 200 classes with diverse styles. ImageNet-A comprises 7,500 real-world adversarial images from 200 categories. CIFAR-100 contains 60,000 $32\times32$ images from 100 classes. DomainNet covers 345 classes across 6 visual domains. Following prior works~\citep{gao2024consistent, he2026task}, we
select the top 200 most populous classes for evaluation.

We follow commonly used evaluation metrics in CL~\citep{tan2024semantically, zhao2024safe, wu2025navigating}: Last Accuracy ($\mathcal{A}_{Last}$), which measures the final performance over all classes, and Average Accuracy ($\mathcal{A}_{Avg}$), which averages the performance across all incremental sessions.

\textbf{Implementation Details.} We use ViT-B/16~\citep{dosovitskiy2021an} pre-trained on ImageNet-21K~\citep{russakovsky2015imagenet} as the backbone. Each task is trained for 20 epochs with a batch size of 48
using DP-SGD with cosine learning-rate decay. The initial learning rate is $0.02$ for ImageNet-R, $0.005$ for CUB, and $0.01$ for the other datasets. We insert \textsf{DP-SPA} into all attention blocks with $r=32$, allocating $r_S=r_P=16$, and train a cosine classifier~\citep{he2026task} alongside the LoRA coefficients. We select the clipping threshold from
$\{0.1,0.5,1,2,5\}$ and use $C=1$.
The main experiments use $\epsilon=3$, and we additionally evaluate $\epsilon\in\{1,3,5,7\}$, with $\delta=10^{-5}$ throughout. We use the RDP accountant implemented in Opacus~\citep{yousefpour2021opacus} to calibrate the Gaussian noise multiplier $\sigma$ to the target $(\epsilon,\delta)$-DP guarantee. All experiments are conducted using NVIDIA RTX 5090 GPUs, and the reported results are averaged over three random seeds.
\vspace{-1mm}
\subsection{Main Results}
\begin{figure*}[h]
    \centering

    \begin{subfigure}{0.33\textwidth}
        \centering
        \includegraphics[width=\linewidth]{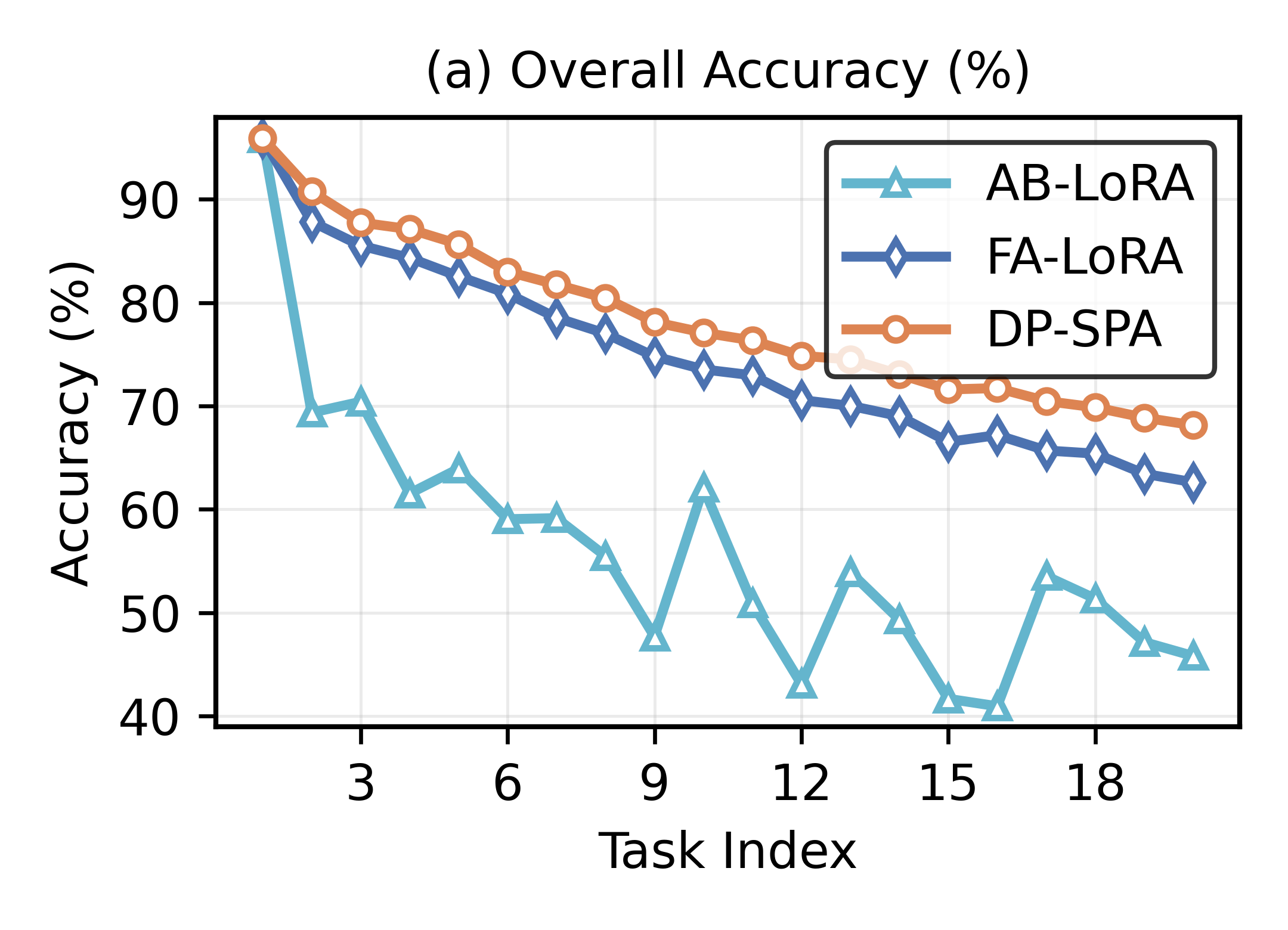}
    \end{subfigure}
    \hspace{-2mm}
    \begin{subfigure}{0.33\textwidth}
        \centering
        \includegraphics[width=\linewidth]{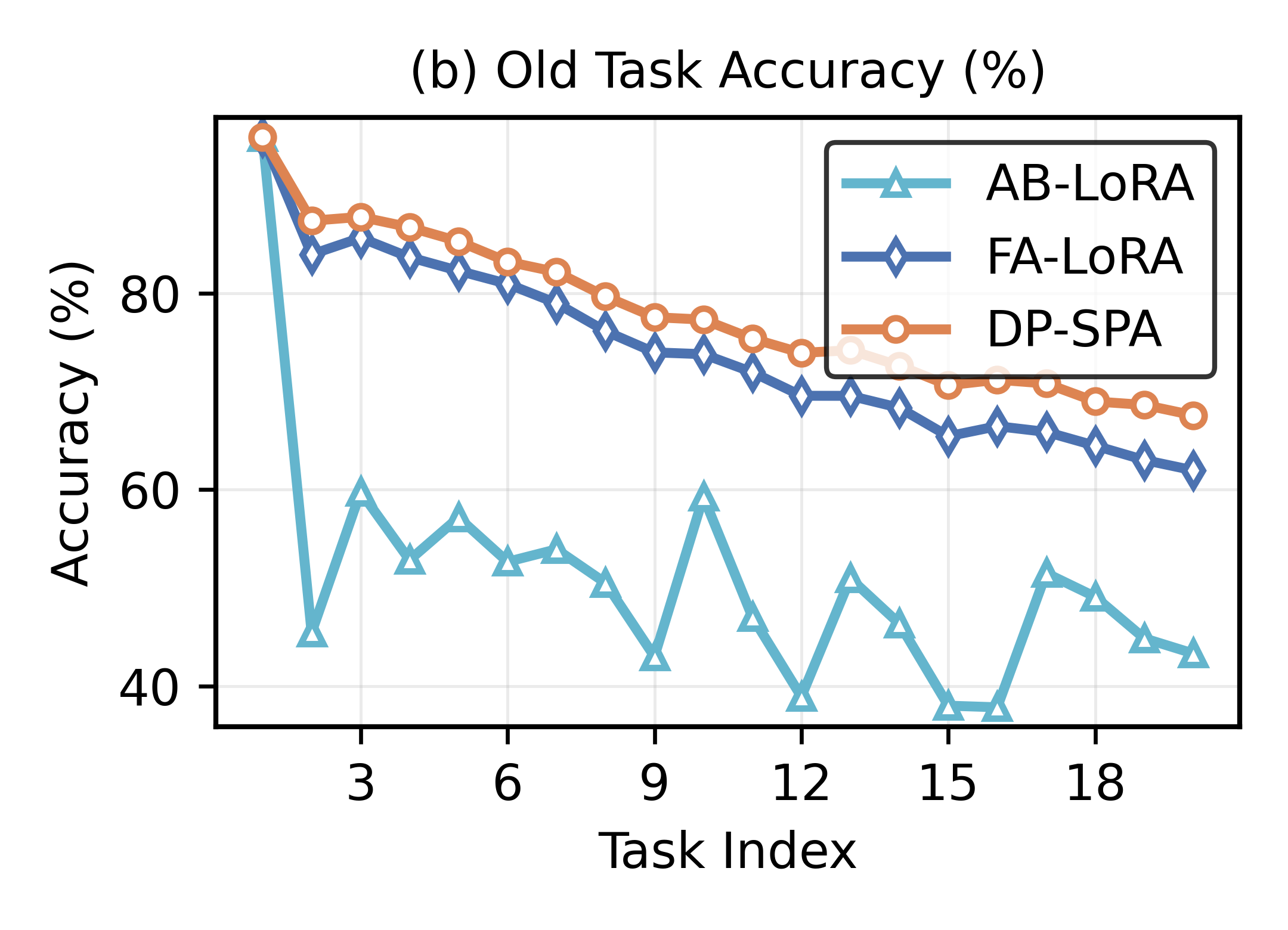}
    \end{subfigure}
    \hspace{-2mm}
    \begin{subfigure}{0.33\textwidth}
        \centering
        \includegraphics[width=\linewidth]{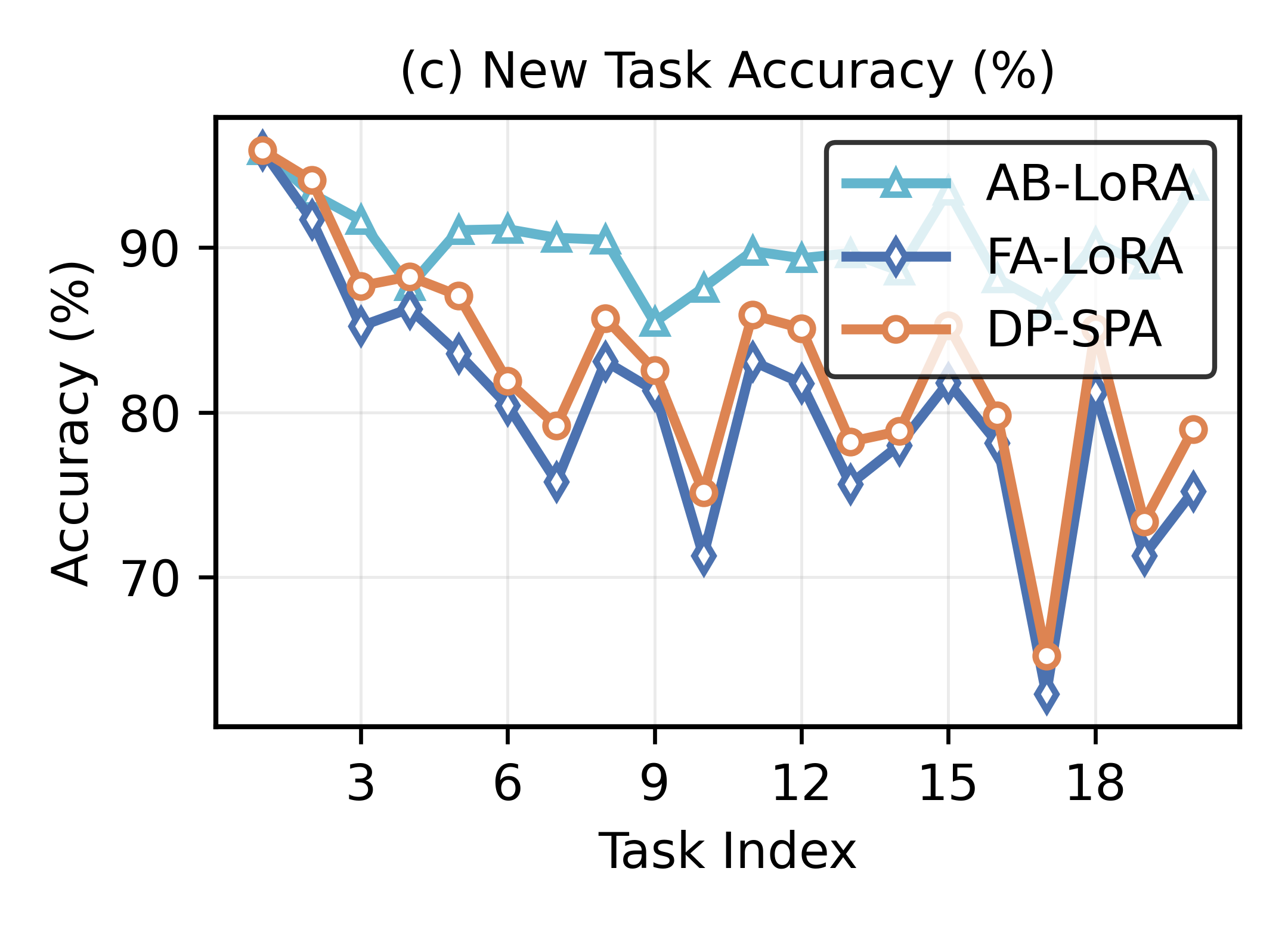}
    \end{subfigure}
    \vspace{-2mm}
    \caption{Overall, old task and new task accuracy on ImageNet-R across 20 incremental tasks.}
    \label{fig:imagenet_r}
\end{figure*}

\textbf{Comparison with Baseline Methods.}
For a controlled comparison under DP, we consider
\textbf{AB-LoRA}, which jointly updates both LoRA factors
$\boldsymbol A$ and $\boldsymbol B$, and a stronger baseline, \textbf{FA-LoRA},
which fixes $\boldsymbol A$ and updates only $\boldsymbol B$.
Tables~\ref{tab:main_results},~\ref{tab:session_results} summarize the results across different datasets and
incremental settings.
Our method substantially outperforms AB-LoRA across all settings.
On ImageNet-R with $T{=}10$, \textsf{DP-SPA} increases
$\mathcal{A}_{Last}$ from $56.86\%$ to $71.85\%$ and
$\mathcal{A}_{Avg}$ from $63.83\%$ to $79.90\%$,
corresponding to gains of $14.99$ and $16.07$ percentage points,
respectively.
The gains increase to $22.31$ and $22.22$ percentage points
for $T{=}20$.
Compared with the stronger FA-LoRA, our method consistently
achieves higher accuracy across all datasets and settings.
On ImageNet-R, \textsf{DP-SPA} improves
$\mathcal{A}_{Last}$ and $\mathcal{A}_{Avg}$ by
$3.60$ and $2.70$ percentage points for $T{=}20$, respectively. Figure~\ref{fig:imagenet_r} shows that \textsf{DP-SPA} maintains higher overall and old-task accuracy while preserving comparable new-task accuracy, demonstrating a better stability--plasticity trade-off.

\begin{table*}[h]
    \centering
    \caption{Experimental results for 10 incremental tasks on ImageNet-R, 
    ImageNet-A, CIFAR-100, and CUB. The best results are highlighted in \textbf{bold}.}
    \label{tab:main_results}

    \setlength{\tabcolsep}{4pt}
    \renewcommand{\arraystretch}{1.15}
    \setlength{\heavyrulewidth}{1.0pt}   
    \setlength{\lightrulewidth}{0.5pt}   
    \setlength{\cmidrulewidth}{0.4pt}    

    \resizebox{\textwidth}{!}{
    \begin{tabular}{lcccccccc}
        \toprule
        \multirow{2}{*}{\textbf{Method}}
        & \multicolumn{2}{c}{\textbf{ImageNet-R} ($T{=}10$)}
        & \multicolumn{2}{c}{\textbf{ImageNet-A} ($T{=}10$)}
        & \multicolumn{2}{c}{\textbf{CIFAR-100} ($T{=}10$)}
        & \multicolumn{2}{c}{\textbf{CUB} ($T{=}10$)} \\
        
        \cmidrule(lr){2-3}
        \cmidrule(lr){4-5}
        \cmidrule(lr){6-7}
        \cmidrule(lr){8-9}

        & $\mathcal{A}_{Last}\uparrow$
        & $\mathcal{A}_{Avg}\uparrow$
        & $\mathcal{A}_{Last}\uparrow$
        & $\mathcal{A}_{Avg}\uparrow$
        & $\mathcal{A}_{Last}\uparrow$
        & $\mathcal{A}_{Avg}\uparrow$
        & $\mathcal{A}_{Last}\uparrow$
        & $\mathcal{A}_{Avg}\uparrow$ \\
        \midrule

        AB-LoRA
        & $56.86_{\pm 1.37}$ & $63.83_{\pm 1.54}$
        & $51.69_{\pm 1.22}$ & $62.23_{\pm 0.92}$
        & $85.42_{\pm 0.36}$ & $90.40_{\pm 0.42}$
        & $72.37_{\pm 0.90}$ & $79.33_{\pm 0.67}$ \\

        FA-LoRA
        & $68.55_{\pm 0.77}$ & $77.46_{\pm 0.89}$
        & $54.01_{\pm 0.78}$ & $64.42_{\pm 0.71}$
        & $86.56_{\pm 0.30}$ & $90.63_{\pm 0.41}$
        & $74.02_{\pm 0.85}$ & $80.26_{\pm 0.71}$ \\

        \textbf{DP-SPA}
        & $\mathbf{71.85_{\pm 0.62}}$
        & $\mathbf{79.90_{\pm 0.57}}$
        & $\mathbf{57.21_{\pm 0.58}}$
        & $\mathbf{66.65_{\pm 0.63}}$
        & $\mathbf{87.07_{\pm 0.22}}$
        & $\mathbf{91.19_{\pm 0.35}}$
        & $\mathbf{75.78_{\pm 0.64}}$
        & $\mathbf{82.67_{\pm 0.59}}$ \\
        \bottomrule
    \end{tabular}
    }
    \vspace{-1mm}
\end{table*}

\begin{table*}[h]
    \centering
    \caption{Experimental results for 5 and 20 incremental tasks on ImageNet-R, 
    20 tasks on ImageNet-A, and 10 tasks on DomainNet. The best results are 
    highlighted in \textbf{bold}.}
    \label{tab:session_results}

    \setlength{\tabcolsep}{4pt}
    \renewcommand{\arraystretch}{1.15}
    \setlength{\heavyrulewidth}{1.0pt}   
    \setlength{\lightrulewidth}{0.5pt}   
    \setlength{\cmidrulewidth}{0.4pt}    

    \resizebox{\textwidth}{!}{
    \begin{tabular}{lcccccccc}
        \toprule
        \multirow{2}{*}{\textbf{Method}}
        & \multicolumn{2}{c}{\textbf{ImageNet-R} ($T{=}5$)}
        & \multicolumn{2}{c}{\textbf{ImageNet-R} ($T{=}20$)}
        & \multicolumn{2}{c}{\textbf{ImageNet-A} ($T{=}20$)}
        & \multicolumn{2}{c}{\textbf{DomainNet} ($T{=}10$)} \\

        \cmidrule(lr){2-3}
        \cmidrule(lr){4-5}
        \cmidrule(lr){6-7}
        \cmidrule(lr){8-9}

        & $\mathcal{A}_{Last}\uparrow$
        & $\mathcal{A}_{Avg}\uparrow$
        & $\mathcal{A}_{Last}\uparrow$
        & $\mathcal{A}_{Avg}\uparrow$
        & $\mathcal{A}_{Last}\uparrow$
        & $\mathcal{A}_{Avg}\uparrow$
        & $\mathcal{A}_{Last}\uparrow$
        & $\mathcal{A}_{Avg}\uparrow$ \\
        \midrule

        AB-LoRA
        & $62.48_{\pm 0.86}$ & $70.73_{\pm 0.74}$
        & $45.83_{\pm 2.96}$ & $56.12_{\pm 2.41}$
        & $41.44_{\pm 2.23}$ & $54.11_{\pm 1.80}$
        & $76.53_{\pm 1.12}$ & $83.22_{\pm 1.05}$ \\

        FA-LoRA
        & $71.66_{\pm 0.59}$ & $79.20_{\pm 0.61}$
        & $64.54_{\pm 0.93}$ & $75.64_{\pm 0.79}$
        & $45.25_{\pm 1.42}$ & $58.12_{\pm 1.04}$
        & $78.07_{\pm 0.64}$ & $84.83_{\pm 0.65}$ \\

        \textbf{DP-SPA}
        & $\mathbf{75.45_{\pm 0.51}}$
        & $\mathbf{82.06_{\pm 0.42}}$
        & $\mathbf{68.14_{\pm 0.75}}$
        & $\mathbf{78.34_{\pm 0.67}}$
        & $\mathbf{48.44_{\pm 0.84}}$
        & $\mathbf{60.89_{\pm 0.78}}$
        & $\mathbf{79.45_{\pm 0.59}}$
        & $\mathbf{85.61_{\pm 0.72}}$ \\
        \bottomrule
    \end{tabular}
    }
\end{table*}

\begin{table*}[h]
    \centering
    \caption{Experimental results under different $\epsilon$ on ImageNet-A and ImageNet-R with $T=20$.}
    \label{tab:privacy_budget}
    \setlength{\tabcolsep}{3.5pt}
    \renewcommand{\arraystretch}{1.15}
    \setlength{\heavyrulewidth}{1.0pt}   
    \setlength{\lightrulewidth}{0.5pt}   
    \setlength{\cmidrulewidth}{0.4pt}    

    \resizebox{\textwidth}{!}{
    \begin{tabular}{llcccccccc}
        \toprule
        \multirow{2}{*}{\textbf{Dataset}}
        & \multirow{2}{*}{\textbf{Method}}
        & \multicolumn{2}{c}{$\boldsymbol{\epsilon=1}$}
        & \multicolumn{2}{c}{$\boldsymbol{\epsilon=3}$}
        & \multicolumn{2}{c}{$\boldsymbol{\epsilon=5}$}
        & \multicolumn{2}{c}{$\boldsymbol{\epsilon=7}$} \\

        \cmidrule(lr){3-4}
        \cmidrule(lr){5-6}
        \cmidrule(lr){7-8}
        \cmidrule(lr){9-10}

        &
        & $\mathcal{A}_{Last}\uparrow$
        & $\mathcal{A}_{Avg}\uparrow$
        & $\mathcal{A}_{Last}\uparrow$
        & $\mathcal{A}_{Avg}\uparrow$
        & $\mathcal{A}_{Last}\uparrow$
        & $\mathcal{A}_{Avg}\uparrow$
        & $\mathcal{A}_{Last}\uparrow$
        & $\mathcal{A}_{Avg}\uparrow$ \\
        \midrule

        \multirow{3}{*}{\textbf{ImageNet-R}}
        & AB-LoRA
        & $3.26_{\pm 4.77}$ & $11.37_{\pm 3.25}$
        & $45.83_{\pm 2.96}$ & $56.12_{\pm 2.41}$
        & $48.73_{\pm 2.66}$ & $57.42_{\pm 1.95}$
        & $50.82_{\pm 1.23}$ & $60.28_{\pm 0.96}$ \\

        & FA-LoRA
        & $34.45_{\pm 2.52}$ & $55.73_{\pm 2.70}$
        & $64.54_{\pm 0.93}$ & $75.64_{\pm 0.79}$
        & $67.25_{\pm 0.88}$ & $78.36_{\pm 0.73}$
        & $70.76_{\pm 0.61}$ & $78.64_{\pm 0.84}$ \\

        & \textbf{DP-SPA}
        & $\mathbf{48.69_{\pm 1.98}}$ & $\mathbf{60.44_{\pm 1.94}}$
        & $\mathbf{68.14_{\pm 0.75}}$ & $\mathbf{78.34_{\pm 0.67}}$
        & $\mathbf{69.85_{\pm 0.70}}$ & $\mathbf{79.65_{\pm 0.72}}$
        & $\mathbf{72.46_{\pm 0.55}}$ & $\mathbf{80.27_{\pm 0.54}}$ \\

        \midrule

        \multirow{3}{*}{\textbf{ImageNet-A}}
        & AB-LoRA
        & $31.03_{\pm 3.35}$ & $45.12_{\pm 2.87}$
        & $41.44_{\pm 2.23}$ & $54.11_{\pm 1.80}$
        & $43.33_{\pm 0.66}$ & $55.08_{\pm 0.73}$
        & $44.08_{\pm 0.68}$ & $55.46_{\pm 0.49}$ \\

        & FA-LoRA
        & $43.30_{\pm 2.95}$ & $56.07_{\pm 2.96}$
        & $45.25_{\pm 1.42}$ & $58.12_{\pm 1.04}$
        & $47.61_{\pm 0.63}$ & $59.75_{\pm 0.75}$
        & $48.24_{\pm 0.49}$ & $60.07_{\pm 0.47}$ \\

        & \textbf{DP-SPA}
        & $\mathbf{45.22_{\pm 1.54}}$ & $\mathbf{58.46_{\pm 1.25}}$
        & $\mathbf{48.44_{\pm 0.84}}$ & $\mathbf{60.89_{\pm 0.78}}$
        & $\mathbf{49.70_{\pm 0.63}}$ & $\mathbf{59.92_{\pm 0.56}}$
        & $\mathbf{50.43_{\pm 0.41}}$ & $\mathbf{61.38_{\pm 0.44}}$ \\
        \bottomrule
    \end{tabular}
    }
    \vspace{-2mm}
\end{table*}

\textbf{Effect of Privacy Budgets.}
Table~\ref{tab:privacy_budget} compares different $\epsilon$ on ImageNet-R and ImageNet-A with $T{=}20$.
\textsf{DP-SPA} achieves the highest mean accuracy across
all evaluated budgets.
At $\epsilon=1$ on ImageNet-R, AB-LoRA achieves only
$3.26\%$ in $\mathcal{A}_{Last}$, whereas FA-LoRA and
\textsf{DP-SPA} achieve $34.45\%$ and $48.69\%$, respectively.
The improvement of FA-LoRA over AB-LoRA supports the
benefit of fixing $\boldsymbol A$ under strong privacy noise.
\textsf{DP-SPA} further outperforms FA-LoRA by $14.24$
percentage points, demonstrating the additional benefit
of its stable--plastic design and recalibration.
As $\epsilon$ increases, this gap narrows, while \textsf{DP-SPA} retains its advantage on both datasets.

\subsection{Ablation Studies.}
\textbf{Effect of Stable and Plastic bases.}
We ablate the stable and plastic bases to examine their individual contributions.
As shown in Table~\ref{tab:ablation_sp}, both
$\boldsymbol{A}_S^t$-only ($(r_S,r_P)=(32,0)$) and $\boldsymbol{A}_P^t$-only ($(0,32)$) consistently
outperform AB-LoRA, with $\boldsymbol{A}_S^t$-only performing better
across all settings.
This advantage becomes more pronounced for longer task sequences,
highlighting the increasing importance of preserving historically
important directions. Combining $\boldsymbol{A}_S^t$ and $\boldsymbol{A}_P^t$ consistently achieves the best performance, demonstrating their complementary roles in balancing stability and plasticity.
\begin{table*}[h]
    \centering
    \caption{Ablation studies on the stable and plastic bases on ImageNet-R and ImageNet-A.}
    \vspace{-1mm}
    \label{tab:ablation_sp}
    \setlength{\tabcolsep}{3.5pt}
    \renewcommand{\arraystretch}{1.15}
    \setlength{\heavyrulewidth}{1.0pt}
    \setlength{\lightrulewidth}{0.5pt}
    \setlength{\cmidrulewidth}{0.4pt}
    \resizebox{\textwidth}{!}{
    \begin{tabular}{lcccccccc}
        \toprule
        \multirow{2}{*}{\textbf{Variant}}
        & \multicolumn{2}{c}{\textbf{ImageNet-R} ($T{=}10$)}
        & \multicolumn{2}{c}{\textbf{ImageNet-R} ($T{=}20$)}
        & \multicolumn{2}{c}{\textbf{ImageNet-A} ($T{=}10$)}
        & \multicolumn{2}{c}{\textbf{ImageNet-A} ($T{=}20$)} \\

        \cmidrule(lr){2-3}
        \cmidrule(lr){4-5}
        \cmidrule(lr){6-7}
        \cmidrule(lr){8-9}

        & $\mathcal{A}_{Last}\uparrow$
        & $\mathcal{A}_{Avg}\uparrow$
        & $\mathcal{A}_{Last}\uparrow$
        & $\mathcal{A}_{Avg}\uparrow$
        & $\mathcal{A}_{Last}\uparrow$
        & $\mathcal{A}_{Avg}\uparrow$
        & $\mathcal{A}_{Last}\uparrow$
        & $\mathcal{A}_{Avg}\uparrow$ \\
        \midrule

        AB-LoRA
        & $56.86_{\pm 1.37}$ & $63.83_{\pm 1.54}$
        & $45.83_{\pm 2.96}$ & $56.12_{\pm 2.41}$
        & $51.69_{\pm 1.22}$ & $62.23_{\pm 0.92}$
        & $41.44_{\pm 2.23}$ & $54.11_{\pm 1.80}$ \\

        $\boldsymbol{A}_S^t$-only
        & $70.53_{\pm 0.36}$ & $78.05_{\pm 0.48}$
        & $67.76_{\pm 0.60}$ & $77.15_{\pm 0.46}$
        & $56.17_{\pm 0.54}$ & $65.13_{\pm 0.70}$
        & $47.05_{\pm 0.65}$ & $59.43_{\pm 0.52}$ \\

        $\boldsymbol{A}_P^t$-only
        & $68.81_{\pm 0.52}$ & $77.94_{\pm 0.57}$
        & $65.80_{\pm 0.74}$ & $76.32_{\pm 0.81}$
        & $54.82_{\pm 0.61}$ & $64.96_{\pm 0.52}$
        & $45.19_{\pm 0.90}$ & $58.24_{\pm 0.68}$ \\

        \textbf{$\boldsymbol{A}_S^t{+}\boldsymbol{A}_P^t$}
        & $\mathbf{71.85_{\pm 0.62}}$
        & $\mathbf{79.90_{\pm 0.57}}$
        & $\mathbf{68.14_{\pm 0.75}}$
        & $\mathbf{78.34_{\pm 0.67}}$
        & $\mathbf{57.21_{\pm 0.58}}$
        & $\mathbf{66.65_{\pm 0.63}}$
        & $\mathbf{48.44_{\pm 0.84}}$
        & $\mathbf{60.89_{\pm 0.78}}$ \\
        \bottomrule
    \end{tabular}
    }
\end{table*}

\begin{wraptable}{r}{0.46\columnwidth}
\small
    \vspace{-12pt}
    \centering
    \caption{Effect of recalibration on ImageNet-R with $T=20$, $r=32$, and $\epsilon=3$.}
    \vspace{-1mm}
    \label{tab:ablation_recalibration}
    \setlength{\tabcolsep}{3.5pt}
    \renewcommand{\arraystretch}{1.15}
    \setlength{\heavyrulewidth}{1.0pt}
    \setlength{\lightrulewidth}{0.5pt}
    \setlength{\cmidrulewidth}{0.4pt}
    \resizebox{0.85\linewidth}{!}{%
        \begin{tabular}{@{}lcc@{}}
            \toprule
            \textbf{Variant}
            & $\mathcal{A}_{Last}\uparrow$
            & $\mathcal{A}_{Avg}\uparrow$ \\
            \midrule
            DP-SPA w/o recal. & $67.25_{\pm 0.69}$ & $77.49_{\pm 0.72}$ \\
            \textbf{DP-SPA}
            & $\mathbf{68.14_{\pm 0.75}}$
            & $\mathbf{78.34_{\pm 0.67}}$ \\
            \bottomrule
        \end{tabular}%
    }
    \vspace{-8pt}
\end{wraptable}

\textbf{Effect of History-Aware Recalibration.}
Table~\ref{tab:ablation_recalibration} isolates recalibration
while retaining the stable--plastic subspace construction.
Recalibration improves $\mathcal{A}_{Last}$ from 67.25\% to
68.14\% and $\mathcal{A}_{Avg}$ from 77.49\% to 78.34\%.
This shows that recalibrating stable-direction updates further improves continual adaptation under privacy constraints.

\section{Discussion and Conclusion}
We study differentially private continual LoRA and observe increased forgetting during subsequent adaptation and lower retention at comparable new-task accuracy in the evaluated settings. Joint-factor privatization also introduces a within-step bilinear noise interaction. \textsf{DP-SPA} constructs a stable basis from privatized updates and a plastic basis from pretrained weights within its orthogonal complement, fixes both bases during private up-projection optimization, and recalibrates stable-direction updates using historical and current update energies. Experiments show improved average accuracy and retention across the evaluated settings and privacy budgets, supporting the reuse of privatized history without additional privacy cost. Recalibration guarantees non-increasing historical overlap; under the stated gradient and curvature assumptions, it also yields a non-increasing upper bound on old-task loss increase, without guaranteeing lower actual loss.

\bibliography{iclr2027_conference}
\bibliographystyle{iclr2027_conference}

%
%

\clearpage
\section{Ethics Statement}
This work adheres to the ICLR Code of Ethics. In this study, no human subjects or animal experimentation was involved. All datasets used, including \textbf{ImageNet-R, ImageNet-A, CIFAR-100, CUB, DomainNet}, were sourced in compliance with relevant usage guidelines, ensuring no violation of privacy. We have taken care to avoid any biases or discriminatory outcomes in our research process. No personally identifiable information was used, and no experiments were conducted that could raise privacy or security concerns. We are committed to maintaining transparency and integrity throughout the research process.

\section{Reproducibility Statement}
We have made every effort to ensure the reproducibility of the results presented in this paper. Detailed experimental settings, including training procedures, model configurations, hyperparameters, and hardware specifications, are provided in Section~\ref{sec:Experimental Setups} and Appendix~\ref{app:setup}. The source code and scripts required to reproduce our experiments will be made publicly available upon acceptance.

\section{AI Use Statement}
\label{LLM_usage}
Large language models (LLMs) were used to assist with
manuscript writing and polishing, including sentence
rephrasing, grammar checking, improving readability and
clarity, and checking the consistency of theoretical
arguments and mathematical notation.

The LLMs were not used to originate the research ideas
or design the methods or experiments. The research
and analyses were conducted by the authors. The authors reviewed and revised all AI-assisted content
to ensure its accuracy and consistency with the underlying
research. We take full responsibility for the final content
of this work, including all text, claims, and artifacts
produced with the assistance of generative AI.

\clearpage
\onecolumn
\appendix
\section*{Appendix Contents}
\begingroup
\small
\setlength{\parindent}{0pt}
\par\medskip\noindent\textbf{\ref{app:table_notations}\quad Table of Notations}\nobreak\hfill\pageref{app:table_notations}\par
\par\medskip\noindent\textbf{\ref{app:algo}\quad DP-SPA Algorithm}\nobreak\hfill\pageref{app:algo}\par
\par\medskip\noindent\textbf{\ref{app:Theoretical}\quad Proofs of the Main Theoretical Results}\nobreak\hfill\pageref{app:Theoretical}\par
\noindent\hspace*{1em}\ref{app:privacy}\quad Privacy guarantee\nobreak\hfill\pageref{app:privacy}\par
\noindent\hspace*{1em}\ref{app:geometry}\quad Stable--plastic geometry\nobreak\hfill\pageref{app:geometry}\par
\noindent\hspace*{1em}\ref{app:recalibration}\quad Recalibration guarantee\nobreak\hfill\pageref{app:recalibration}\par
\noindent\hspace*{1em}\ref{app:forgetting}\quad Conditional old-task loss bound\nobreak\hfill\pageref{app:forgetting}\par
\par\medskip\noindent\textbf{\ref{app:noise}\quad Supplementary Analysis of the Bilinear Noise Interaction}\nobreak\hfill\pageref{app:noise}\par
\par\medskip\noindent\textbf{\ref{app:experimental}\quad Supplementary Experimental Information}\nobreak\hfill\pageref{app:experimental}\par
\noindent\hspace*{1em}\ref{app:setup}\quad Experimental Details\nobreak\hfill\pageref{app:setup}\par
\noindent\hspace*{1em}\ref{app:Additional Results}\quad Additional Results\nobreak\hfill\pageref{app:Additional Results}\par
\endgroup
\clearpage

\section{Table of Notations}
\label{app:table_notations}

Table~\ref{tab:notations} summarizes the main notations used throughout the paper.


\begin{table}[h]
    \centering
    \caption{Summary of the main notations.}
    \label{tab:notations}
    \setlength{\tabcolsep}{10pt}
    \renewcommand{\arraystretch}{1.12}
    \begin{tabular}{lp{0.68\linewidth}}
        \toprule
        \textbf{Notation} & \textbf{Description} \\
        \midrule
        $T,\ t$ & Number of tasks and current task index \\
        $\mathcal{D}^t,\ \mathcal{C}^t$ &
        Private dataset and class set of task $t$ \\
        $\boldsymbol W^0\in\mathbb{R}^{m\times n}$ &
        Pretrained weight matrix \\
        $\boldsymbol W^t$ &
        Weight matrix after completing task $t$ \\
        $r_S,\ r_P,\ r=r_S+r_P$ &
        Stable, plastic, and total LoRA ranks \\
        $\mathcal{S}^t,\ \mathcal{P}^t$ &
        Stable and plastic input-side subspaces \\
        $\boldsymbol A_S^t,\ \boldsymbol A_P^t$ &
        Orthonormal row bases of $\mathcal{S}^t$ and $\mathcal{P}^t$ \\
        $\boldsymbol A^t=
        [(\boldsymbol A_S^t)^\top,
        (\boldsymbol A_P^t)^\top]^\top$ &
        Fixed LoRA basis for task $t$ \\
        $\boldsymbol B^t=[\boldsymbol B_S^t,\boldsymbol B_P^t]$ &
        Privately trained LoRA coefficient matrix \\
        $\boldsymbol H^{t-1}$ &
        Historical statistic of previously merged updates \\
        $\beta$ &
        Decay factor for historical updates \\
        $h_j^t$&
        Eigenvalue $h_j^t$ of the history matrix\\
         $c_j^t$ &
         current stable-column energy $c_j^t$ \\
        $\lambda_j^t,\ \boldsymbol\Lambda_S^t$ &
        Direction-wise recalibration coefficient and its diagonal matrix \\
        $\Delta\boldsymbol W_{\mathrm{priv}}^t$ &
        Private task update before recalibration \\
        $\Delta\boldsymbol W^t$ &
        Merged task update; recalibration is omitted at $t=1$ \\
        $\mathcal I_t(\boldsymbol U)$ &
        Historical overlap energy of an update $\boldsymbol U$ \\
        $(\epsilon_t,\delta_t),\ (\epsilon,\delta)$ &
Task-specific and common record-level privacy budgets \\
        $C,\ \sigma$ &
        Per-record clipping norm and Gaussian noise multiplier \\
        $\mathcal{A}_{Last},\ \mathcal{A}_{Avg}$ &
        Last Accuracy and Average Accuracy \\
        $\tau$ &
Regularization constant in history-aware recalibration \\

$\boldsymbol C_P^t,\ \mu_j^t$ &
Restricted pretrained Gram matrix and its $j$-th eigenvalue \\
        \bottomrule
    \end{tabular}
\end{table}

\section{DP-SPA Algorithm}
\label{app:algo}
Algorithm~\ref{alg:dp_spa} summarizes the complete training procedure of \textsf{DP-SPA}, including stable--plastic basis construction, private LoRA optimization, history-aware recalibration, and model merging.

\begin{algorithm}[ht]
\caption{Differentially Private Stable-Plastic Low-rank Adaptation
(\textsf{DP-SPA})}
\label{alg:dp_spa}
\begin{algorithmic}[1]

\State \textbf{Input:} Pretrained model $\boldsymbol W^0$,
private task sequence $\{\mathcal D^t\}_{t=1}^{T}$,
ranks $r_S,r_P$ with $r=r_S+r_P$,
DP parameters $(C,\sigma)$, learning rate $\eta$,
decay $\beta$, regularization constant $\tau>0$

\State \textbf{Output:} Updated continual model
$\boldsymbol W^T$ with merged LoRA updates

\State Initialize historical statistic
$\boldsymbol H^0\gets\boldsymbol 0$

\For{task $t=1$ to $T$}
    \If{$t=1$}
        \State Initialize $\boldsymbol A^1$ from $\boldsymbol W^0$:
    $\boldsymbol W^0=\boldsymbol U_0\boldsymbol\Sigma_0\boldsymbol V_0^\top$,
    $\boldsymbol A^1\gets\boldsymbol V_{0,1:r}^\top$
    \Else
        \State Construct stable basis:
        $\boldsymbol H^{t-1}=
        \boldsymbol V_H^t\boldsymbol\Lambda_H^t
        (\boldsymbol V_H^t)^\top$,
        $\boldsymbol A_S^t\gets
        (\boldsymbol V_{H,1:r_S}^t)^\top$

        \State Compute
        $\boldsymbol C_P^t\gets
        (\boldsymbol U_\perp^t)^\top
        (\boldsymbol W^0)^\top\boldsymbol W^0
        \boldsymbol U_\perp^t$

        \State Construct plastic basis:
        $\boldsymbol A_P^t\gets
        (\boldsymbol Z_{P,1:r_P}^t)^\top
        (\boldsymbol U_\perp^t)^\top$

        \State $\boldsymbol A^t\gets
        [(\boldsymbol A_S^t)^\top,
         (\boldsymbol A_P^t)^\top]^\top$
    \EndIf

    \State Initialize $\boldsymbol B^t\gets\boldsymbol 0$,
    freeze $\boldsymbol A^t$, and privately optimize
    $\boldsymbol B^t$ on $\mathcal D^t$ with DP-SGD

    \If{$t=1$}
        \State $\Delta\boldsymbol W^1\gets
        \boldsymbol B^1\boldsymbol A^1$
    \Else
        \State Decompose
        $\boldsymbol B^t=
        [\boldsymbol B_S^t,\boldsymbol B_P^t]$

        \State Compute
        $h_j^t=
        \boldsymbol a_{S,j}^t\boldsymbol H^{t-1}
        (\boldsymbol a_{S,j}^t)^\top$
        and $c_j^t=\|\boldsymbol b_{S,j}^t\|_2^2$
        for $j=1,\ldots,r_S$

        \State Compute
        $\lambda_j^t={c_j^t}/({h_j^t+c_j^t+\tau})$
        for $j=1,\ldots,r_S$ and set
        $\boldsymbol\Lambda_S^t\gets
        \operatorname{diag}(\lambda_1^t,\ldots,\lambda_{r_S}^t)$

        \State $\Delta\boldsymbol W^t\gets
        \boldsymbol B_S^t\boldsymbol\Lambda_S^t
        \boldsymbol A_S^t+
        \boldsymbol B_P^t\boldsymbol A_P^t$
    \EndIf

    \State Integrate the LoRA update:
    $\boldsymbol W^t\gets
    \boldsymbol W^{t-1}+\Delta\boldsymbol W^t$

    \State Update the historical statistic:
    $\boldsymbol H^t\gets
    \beta\boldsymbol H^{t-1}
    +(\Delta\boldsymbol W^t)^\top\Delta\boldsymbol W^t$
\EndFor

\end{algorithmic}
\end{algorithm}

\section{Proofs of the Main Theoretical Results}
\label{app:Theoretical}
The following four subsections prove the corresponding main-text theorems.
The final subsection analyzes the fresh-noise interaction in Observation~2.
Unless stated otherwise, geometric and recalibration results concern one
adapted matrix at a fixed task $t\ge2$, with
$0\le r_S$, $0\le r_P$, and $r_S+r_P=r\le n$.
Define
\[
\boldsymbol P^t=(\boldsymbol A^t)^\top\boldsymbol A^t,
\qquad
\boldsymbol Q_S^t=\boldsymbol I_n-(\boldsymbol A_S^t)^\top\boldsymbol A_S^t.
\]
The history matrix is positive semidefinite, with ordered eigenvalues
$h_1^t\ge\cdots\ge h_n^t\ge0$. Its associated overlap functional is
\begin{equation}
\mathcal I_t(\boldsymbol U)
=\operatorname{tr}(\boldsymbol U\boldsymbol H^{t-1}\boldsymbol U^\top)
=\sum_{i<t}\beta^{t-1-i}\|\boldsymbol U(\Delta\boldsymbol W^i)^\top\|_F^2.
\label{eq:hist-proxy}
\end{equation}
The second identity follows by substituting the definition of
$\boldsymbol H^{t-1}$ and taking traces. This functional measures weighted
overlap with historical updates; a connection to old-task loss requires the
assumptions in Theorem~\ref{thm:forgetting_main}.

\subsection{Proof of the Privacy Guarantee}
\label{app:privacy}

\paragraph{Proof of Theorem~\ref{thm:dp_main}.}
Let
$\mathcal D=(\mathcal D^1,\ldots,\mathcal D^T)$, $\mathcal D'=(\mathcal D'^1,\ldots,\mathcal D'^T)$
be neighboring continual datasets under the stated
add/remove adjacency.
Since each protected record belongs to exactly one fixed
task partition, there is a task $s$ such that
$\mathcal D^s$ and $\mathcal D'^s$ differ by one record,
while $\mathcal D^t=\mathcal D'^t$ for all $t\ne s$.

Let $Z_t$ denote the privatized outputs of task $t$,
including the privatized information used to construct
any state passed to subsequent tasks, and write
$Z_{<s}=(Z_1,\ldots,Z_{s-1})$.
Under the theorem's assumptions, all remaining carried
state is public or computed from these privatized outputs.
Tasks preceding $s$ access identical datasets under
$\mathcal D$ and $\mathcal D'$, so $Z_{<s}$ has the same
distribution under both sequences.
Denote this common distribution by $\mu$.

For a fixed prefix $z$, let
$M_s(\mathcal D^s;z)$ denote the task-$s$ mechanism.
By assumption, this mechanism is
$(\epsilon_s,\delta_s)$-DP with respect to
$\mathcal D^s$ for every preceding private transcript $z$.
For any measurable event $E$ in the joint output space of
$(Z_{<s},Z_s)$, define
$E_z=\{y:(z,y)\in E\}$.
Applying the conditional DP inequality and integrating
against the common prefix distribution gives
\begin{align*}
&\Pr_{\mathcal D}\bigl[(Z_{<s},Z_s)\in E\bigr]\\
&\quad=
\int
\Pr\bigl[M_s(\mathcal D^s;z)\in E_z\bigr]\,\mu(dz)\\
&\quad\le
\int
\left(
e^{\epsilon_s}
\Pr\bigl[M_s(\mathcal D'^s;z)\in E_z\bigr]
+\delta_s
\right)\mu(dz)\\
&\quad=
e^{\epsilon_s}
\Pr_{\mathcal D'}\bigl[(Z_{<s},Z_s)\in E\bigr]
+\delta_s.
\end{align*}
Thus, the joint transcript through task $s$ satisfies
the $(\epsilon_s,\delta_s)$ privacy inequality for
this neighboring pair.

For every subsequent task $t>s$,
$\mathcal D^t=\mathcal D'^t$.
The remaining computation accesses only these unchanged
datasets, public information, and the preceding privatized
outputs. Therefore, for the fixed neighboring pair,
it is the same randomized post-processing of
$(Z_{<s},Z_s)$ under both sequences, with the later
datasets treated as fixed auxiliary inputs.
The post-processing property preserves the
$(\epsilon_s,\delta_s)$ inequality for the complete
transcript, including all released models and derived
statistics.
The same argument applies with $\mathcal D$ and
$\mathcal D'$ interchanged.

Because the differing record may belong to any task $s$,
the complete continual mechanism satisfies
\[
\left(
\max_{1\le t\le T}\epsilon_t,\,
\max_{1\le t\le T}\delta_t
\right)\text{-DP}.
\]
In particular, assigning every task the same
$(\epsilon,\delta)$ budget gives an
$(\epsilon,\delta)$ guarantee for the entire sequence.
\hfill$\square$\par

\paragraph{Application to \textsf{DP-SPA}.}
The stable basis is constructed from
$\boldsymbol H^{t-1}$, which depends only on previously
privatized and merged updates.
The plastic basis is constructed from this stable basis
and the public pretrained weights $\boldsymbol W^0$.
After private optimization, the current energies
$c_j^t=\|\boldsymbol b_{S,j}^t\|_2^2$ are computed from
the privatized coefficients $\boldsymbol B^t$,
while the historical energies $h_j^t$ are computed from
$\boldsymbol H^{t-1}$.
Consequently, the recalibration coefficients
$\lambda_j^t$, the merged update
$\Delta\boldsymbol W^t$, the model $\boldsymbol W^t$,
and the updated history $\boldsymbol H^t$ are all
functions of public information and privatized outputs.
These operations therefore incur no additional privacy
cost beyond that of the task-level private optimizer.

\paragraph{Scope.}
The result permits adaptive dependence on earlier
privatized outputs; it does not require independent
models or task mechanisms.
Its record-level guarantee relies on each protected
record belonging to only one task and on subsequent
tasks accessing that record only through privatized
outputs.
It does not directly provide user-level privacy when
one user contributes multiple records across tasks.

\subsection{Proof of the Stable--Plastic Geometry}
\label{app:geometry}

\paragraph{Proof of Theorem~\ref{thm:geometry_main}.}
Fix a task $t\ge2$, with nonnegative integer ranks
$r_S,r_P$ satisfying $r_S+r_P=r\le n$.
We use the bases
$\boldsymbol A_S^t\in\mathbb R^{r_S\times n}$ and
$\boldsymbol A_P^t\in\mathbb R^{r_P\times n}$
constructed in the main text.
Define
\[
\boldsymbol P_S^t
=(\boldsymbol A_S^t)^\top\boldsymbol A_S^t,
\qquad
\boldsymbol P_P^t
=(\boldsymbol A_P^t)^\top\boldsymbol A_P^t,
\qquad
\boldsymbol Q_S^t=\boldsymbol I_n-\boldsymbol P_S^t.
\]
For an update $\boldsymbol U\in\mathbb R^{m\times n}$,
recall that
$  
\mathcal I_t(\boldsymbol U)
=\operatorname{tr}\!\left(
\boldsymbol U\boldsymbol H^{t-1}\boldsymbol U^\top
\right).
$

The stable basis consists of orthonormal eigenvectors,
so
$\boldsymbol A_S^t(\boldsymbol A_S^t)^\top
=\boldsymbol I_{r_S}$.
Let
$\boldsymbol U_\perp^t\in\mathbb R^{n\times(n-r_S)}$
have orthonormal columns spanning $(\mathcal S^t)^\perp$.
Then
\[
(\boldsymbol U_\perp^t)^\top\boldsymbol U_\perp^t
=\boldsymbol I_{n-r_S},
\qquad
\boldsymbol U_\perp^t(\boldsymbol U_\perp^t)^\top
=\boldsymbol Q_S^t.
\]
The plastic construction uses
\[
\boldsymbol C_P^t
=(\boldsymbol U_\perp^t)^\top
(\boldsymbol W^0)^\top\boldsymbol W^0
\boldsymbol U_\perp^t,
\qquad
\boldsymbol A_P^t
=(\boldsymbol Z_{P,1:r_P}^t)^\top
(\boldsymbol U_\perp^t)^\top,
\]
where
$\boldsymbol Z_{P,1:r_P}^t
\in\mathbb R^{(n-r_S)\times r_P}$
contains orthonormal eigenvectors associated with the
$r_P$ largest eigenvalues of $\boldsymbol C_P^t$.
Therefore,
$ 
\boldsymbol A_P^t(\boldsymbol A_P^t)^\top
=\boldsymbol I_{r_P},
\qquad
\boldsymbol A_P^t(\boldsymbol A_S^t)^\top
=\boldsymbol0.
$ 
It follows that
\[
\boldsymbol A^t(\boldsymbol A^t)^\top
=\boldsymbol I_r,
\qquad
\boldsymbol P^t
:=(\boldsymbol A^t)^\top\boldsymbol A^t
=\boldsymbol P_S^t+\boldsymbol P_P^t.
\]
In particular, $\boldsymbol P_S^t$ and
$\boldsymbol P_P^t$ are orthogonal projectors with
$\boldsymbol P_S^t\boldsymbol P_P^t=\boldsymbol0$.

\paragraph{(i) Historical energy capture.}
Write the eigendecomposition
\[
\boldsymbol H^{t-1}
=\boldsymbol V_H^t
\operatorname{diag}(h_1^t,\ldots,h_n^t)
(\boldsymbol V_H^t)^\top,
\qquad
h_1^t\ge\cdots\ge h_n^t\ge0.
\]
Let $\boldsymbol v_{H,j}^t$ denote the $j$-th column of
$\boldsymbol V_H^t$.
Choose this orthonormal eigenbasis consistently with
the stable construction, so that
$\boldsymbol A_S^t=(\boldsymbol V_{H,1:r_S}^t)^\top$,
including when eigenvalues are repeated.

For any $\boldsymbol U\in\mathbb R^{r_S\times n}$
satisfying
$\boldsymbol U\boldsymbol U^\top=\boldsymbol I_{r_S}$,
\[
\operatorname{tr}\!\left(
\boldsymbol U\boldsymbol H^{t-1}\boldsymbol U^\top
\right)
=\sum_{j=1}^{n}h_j^td_j,
\qquad
d_j:=\|\boldsymbol U\boldsymbol v_{H,j}^t\|_2^2.
\]
Since $\boldsymbol U^\top\boldsymbol U$ is an
orthogonal projector of rank $r_S$,
$ 
0\le d_j\le1,
\sum_{j=1}^{n}d_j
=\operatorname{tr}(\boldsymbol U^\top\boldsymbol U)
=r_S.
$ 
For $1\le r_S<n$, these relations imply
\begin{align*}
\sum_{j=1}^{n}h_j^td_j
&\le
\sum_{j=1}^{r_S}h_j^td_j
+h_{r_S}^t\sum_{j=r_S+1}^{n}d_j\\
&=
\sum_{j=1}^{r_S}
\left[h_j^td_j+h_{r_S}^t(1-d_j)\right]\\
&\le \sum_{j=1}^{r_S}h_j^t.
\end{align*}
The cases $r_S=0$ and $r_S=n$ follow directly.
The selected stable basis attains this upper bound,
and hence, for $\boldsymbol U\in\mathbb R^{r_S\times n}$,
\[
\max_{\boldsymbol U}\quad
\operatorname{tr}\!\left(
\boldsymbol U\boldsymbol H^{t-1}\boldsymbol U^\top
\right)
=
\operatorname{tr}\!\left(
\boldsymbol A_S^t\boldsymbol H^{t-1}
(\boldsymbol A_S^t)^\top
\right)
=
\sum_{j=1}^{r_S}h_j^t,\quad
\text{s.t.}\quad
\boldsymbol U\boldsymbol U^\top=\boldsymbol I_{r_S}.
\]

For $r_S<n$, the complementary projector satisfies
\[
\boldsymbol Q_S^t\boldsymbol H^{t-1}\boldsymbol Q_S^t
=
\sum_{j=r_S+1}^{n}
h_j^t\boldsymbol v_{H,j}^t
(\boldsymbol v_{H,j}^t)^\top
\preceq h_{r_S+1}^t\boldsymbol Q_S^t.
\]
Because
$\boldsymbol A_P^t\boldsymbol Q_S^t=\boldsymbol A_P^t$,
\[
\boldsymbol A_P^t\boldsymbol H^{t-1}
(\boldsymbol A_P^t)^\top
\preceq h_{r_S+1}^t\boldsymbol I_{r_P}.
\]
Consequently, for every
$\boldsymbol B_P^t\in\mathbb R^{m\times r_P}$,
\begin{align*}
\mathcal I_t(\boldsymbol B_P^t\boldsymbol A_P^t)
&=
\operatorname{tr}\!\left(
\boldsymbol B_P^t
\boldsymbol A_P^t\boldsymbol H^{t-1}
(\boldsymbol A_P^t)^\top
(\boldsymbol B_P^t)^\top
\right)\\
&\le
h_{r_S+1}^t
\operatorname{tr}\!\left(
\boldsymbol B_P^t(\boldsymbol B_P^t)^\top
\right)\\
&=
h_{r_S+1}^t\|\boldsymbol B_P^t\|_F^2.
\end{align*}
If $r_S=n$, then $r_P=0$ and the plastic update is
zero; the overlap bound is interpreted as zero
without introducing $h_{n+1}^t$.

\paragraph{(ii) Complementary pretrained response.}
Let
$\mu_1^t\ge\cdots\ge\mu_{n-r_S}^t\ge0$
be the eigenvalues of $\boldsymbol C_P^t$.
Consider any
$\boldsymbol U\in\mathbb R^{r_P\times n}$
satisfying
\[
\boldsymbol U\boldsymbol U^\top=\boldsymbol I_{r_P},
\qquad
\boldsymbol U(\boldsymbol A_S^t)^\top=\boldsymbol0.
\]
Set
\[
\widetilde{\boldsymbol U}
:=\boldsymbol U\boldsymbol U_\perp^t
\in\mathbb R^{r_P\times(n-r_S)}.
\]
The orthogonality constraint gives
$\boldsymbol U\boldsymbol Q_S^t=\boldsymbol U$,
and therefore
\[
\boldsymbol U
=\widetilde{\boldsymbol U}(\boldsymbol U_\perp^t)^\top,
\qquad
\widetilde{\boldsymbol U}
\widetilde{\boldsymbol U}^\top
=
\boldsymbol U\boldsymbol Q_S^t\boldsymbol U^\top
=\boldsymbol I_{r_P}.
\]
Conversely, every row-orthonormal
$\widetilde{\boldsymbol U}$ of this dimension produces
an admissible $\boldsymbol U$ through this expression.
Thus,
\begin{align*}
\|\boldsymbol W^0\boldsymbol U^\top\|_F^2
&=
\operatorname{tr}\!\left(
\boldsymbol U(\boldsymbol W^0)^\top
\boldsymbol W^0\boldsymbol U^\top
\right)\\
&=
\operatorname{tr}\!\left(
\widetilde{\boldsymbol U}\boldsymbol C_P^t
\widetilde{\boldsymbol U}^\top
\right)\\
&\le \sum_{j=1}^{r_P}\mu_j^t,
\end{align*}
where the last step applies the eigenvalue argument
from part (i) to $\boldsymbol C_P^t$.
Equality holds for
$\widetilde{\boldsymbol U}
=(\boldsymbol Z_{P,1:r_P}^t)^\top$,
which yields the selected plastic basis.
Hence
for $\boldsymbol U\in\mathbb R^{r_P\times n}$,
\[
\max_{\boldsymbol U}
\|\boldsymbol W^0\boldsymbol U^\top\|_F^2
=
\|\boldsymbol W^0(\boldsymbol A_P^t)^\top\|_F^2
=
\sum_{j=1}^{r_P}\mu_j^t,
\quad
\text{s.t.}\;
\boldsymbol U\boldsymbol U^\top=\boldsymbol I_{r_P},\;
\boldsymbol U(\boldsymbol A_S^t)^\top=\boldsymbol0.
\]
Repeated or zero eigenvalues do not change this
optimal value: an orthonormal selection of the
corresponding eigenvectors in the restricted space
still attains it.

\paragraph{(iii) Additional adaptation capacity.}
For every $\boldsymbol Y\in\mathbb R^{m\times r}$,
row orthonormality gives
\[
\|\boldsymbol Y\boldsymbol A^t\|_F^2
=
\operatorname{tr}\!\left(
\boldsymbol Y\boldsymbol A^t
(\boldsymbol A^t)^\top\boldsymbol Y^\top
\right)
=\|\boldsymbol Y\|_F^2.
\]
This proves the claimed norm identity.

For a target update
$\boldsymbol U\in\mathbb R^{m\times n}$
and coefficients
$\boldsymbol X\in\mathbb R^{m\times r}$,
decompose
\[
\boldsymbol U-\boldsymbol X\boldsymbol A^t
=
\boldsymbol U(\boldsymbol I_n-\boldsymbol P^t)
+
\left(\boldsymbol U(\boldsymbol A^t)^\top
-\boldsymbol X\right)\boldsymbol A^t.
\]
The two terms are orthogonal in the Frobenius inner
product because
$(\boldsymbol I_n-\boldsymbol P^t)
(\boldsymbol A^t)^\top=\boldsymbol0$.
Applying the norm identity gives
\[
\|\boldsymbol U-\boldsymbol X\boldsymbol A^t\|_F^2
=
\|\boldsymbol U(\boldsymbol I_n-\boldsymbol P^t)\|_F^2
+
\|\boldsymbol U(\boldsymbol A^t)^\top-\boldsymbol X\|_F^2.
\]
The unique minimizing coefficient is therefore
$\boldsymbol X=\boldsymbol U(\boldsymbol A^t)^\top$,
with
\[
\min_{\boldsymbol X\in\mathbb R^{m\times r}}
\|\boldsymbol U-\boldsymbol X\boldsymbol A^t\|_F^2
=
\|\boldsymbol U(\boldsymbol I_n-\boldsymbol P^t)\|_F^2.
\]
The same argument applied to the stable basis yields
\[
\min_{\boldsymbol X_S\in\mathbb R^{m\times r_S}}
\|\boldsymbol U-\boldsymbol X_S\boldsymbol A_S^t\|_F^2
=
\|\boldsymbol U(\boldsymbol I_n-\boldsymbol P_S^t)\|_F^2.
\]

Since
$\boldsymbol I_n-\boldsymbol P_S^t
=(\boldsymbol I_n-\boldsymbol P^t)+\boldsymbol P_P^t$
is a sum of mutually orthogonal projectors,
\[
\|\boldsymbol U(\boldsymbol I_n-\boldsymbol P_S^t)\|_F^2
=
\|\boldsymbol U(\boldsymbol I_n-\boldsymbol P^t)\|_F^2
+\|\boldsymbol U\boldsymbol P_P^t\|_F^2.
\]
Moreover,
\[
\|\boldsymbol U\boldsymbol P_P^t\|_F^2
=
\|(\boldsymbol U(\boldsymbol A_P^t)^\top)
\boldsymbol A_P^t\|_F^2
=
\|\boldsymbol U(\boldsymbol A_P^t)^\top\|_F^2,
\]
using row orthonormality of $\boldsymbol A_P^t$.
Subtracting the optimal residuals gives
\[
\min_{\boldsymbol X_S\in\mathbb R^{m\times r_S}}
\|\boldsymbol U-\boldsymbol X_S\boldsymbol A_S^t\|_F^2
-
\min_{\boldsymbol X\in\mathbb R^{m\times r}}
\|\boldsymbol U-\boldsymbol X\boldsymbol A^t\|_F^2
=
\|\boldsymbol U(\boldsymbol A_P^t)^\top\|_F^2.
\]
This establishes all claims of the theorem.
Empty branches are interpreted using empty matrices
and zero sums.
\hfill$\square$\par

\paragraph{Interpretation.}
The approximation identity compares the same stable
basis with and without the selected plastic branch.
It quantifies the additional representational capacity
provided by that branch, rather than superiority over
all alternative bases of total rank $r$.

\subsection{Proof of the Recalibration Guarantee}
\label{app:recalibration}

\paragraph{Proof of Theorem~\ref{thm:recalibration_main}.}
Fix a task $t\ge2$ and condition on the realized private
coefficient matrix
$\boldsymbol B^t=[\boldsymbol B_S^t,\boldsymbol B_P^t]$
and history $\boldsymbol H^{t-1}$.
For each stable direction $j$, let
$\boldsymbol b_{S,j}^t\in\mathbb R^m$ denote the
$j$-th column of $\boldsymbol B_S^t$, and recall
$ 
c_j^t=\|\boldsymbol b_{S,j}^t\|_2^2,
\lambda_j^t=\frac{c_j^t}{h_j^t+c_j^t+\tau},
\boldsymbol\Lambda_S^t
=\operatorname{diag}(\lambda_1^t,\ldots,\lambda_{r_S}^t).
$ 
Here $h_j^t\ge0$ is the historical eigenvalue associated
with the $j$-th stable basis vector, and $\tau>0$.
Thus $0\le\lambda_j^t<1$.

Define the recalibrated coefficient matrix
$
\widehat{\boldsymbol B}^t
=
[\boldsymbol B_S^t\boldsymbol\Lambda_S^t,
 \boldsymbol B_P^t].
$ 
The updates before and after recalibration are
$\Delta\boldsymbol W_{\mathrm{priv}}^t
=\boldsymbol B^t\boldsymbol A^t$
and
$\Delta\boldsymbol W^t
=\widehat{\boldsymbol B}^t\boldsymbol A^t$,
respectively.

\paragraph{(i) Unique minimizer of the recalibration objective.}
Write
$\boldsymbol X=[\boldsymbol X_S,\boldsymbol X_P]$,
where
$\boldsymbol X_S\in\mathbb R^{m\times r_S}$ and
$\boldsymbol X_P\in\mathbb R^{m\times r_P}$,
and let $\boldsymbol x_j$ denote the $j$-th column
of $\boldsymbol X_S$.
The objective in the theorem is
\[
J_t(\boldsymbol X)
=
\frac12\|\boldsymbol X-\boldsymbol B^t\|_F^2
+
\frac12\sum_{j:c_j^t>0}
\frac{h_j^t+\tau}{c_j^t}\|\boldsymbol x_j\|_2^2,
\]
subject to $\boldsymbol x_j=\boldsymbol0$ whenever
$c_j^t=0$.
Because the Frobenius norm separates over columns,
each free stable column minimizes
\[
J_{t,j}(\boldsymbol x_j)
=
\frac12\|\boldsymbol x_j-\boldsymbol b_{S,j}^t\|_2^2
+
\frac{h_j^t+\tau}{2c_j^t}\|\boldsymbol x_j\|_2^2.
\]
For $c_j^t>0$, its gradient and Hessian are
\[
\nabla J_{t,j}(\boldsymbol x_j)
=
\left(1+\frac{h_j^t+\tau}{c_j^t}\right)
\boldsymbol x_j-\boldsymbol b_{S,j}^t,
\qquad
\nabla^2J_{t,j}
=
\left(1+\frac{h_j^t+\tau}{c_j^t}\right)
\boldsymbol I_m\succ0.
\]
Setting the gradient to zero therefore gives the
unique minimizer
\[
\boldsymbol x_j
=
\frac{c_j^t}{c_j^t+h_j^t+\tau}
\boldsymbol b_{S,j}^t
=
\lambda_j^t\boldsymbol b_{S,j}^t.
\]
If $c_j^t=0$, then
$\boldsymbol b_{S,j}^t=\boldsymbol0$ and
$\lambda_j^t=0$, so the imposed constraint
$\boldsymbol x_j=\boldsymbol0$ agrees with the same rule.

The plastic block contributes only
$\frac12\|\boldsymbol X_P-\boldsymbol B_P^t\|_F^2$,
whose unique minimizer is
$\boldsymbol X_P=\boldsymbol B_P^t$.
Combining these columnwise solutions proves that
$\widehat{\boldsymbol B}^t$ is the unique minimizer
of the full constrained objective.

\paragraph{(ii) Plastic coefficients and update norms.}
The orthogonality relations established in
Theorem~\ref{thm:geometry_main} give
$ 
\boldsymbol A_S^t(\boldsymbol A_P^t)^\top
=\boldsymbol0,
\qquad
\boldsymbol A_P^t(\boldsymbol A_P^t)^\top
=\boldsymbol I_{r_P}.
$ 
Consequently,
\begin{align*}
\Delta\boldsymbol W^t(\boldsymbol A_P^t)^\top
&=
\boldsymbol B_S^t\boldsymbol\Lambda_S^t
\boldsymbol A_S^t(\boldsymbol A_P^t)^\top
+
\boldsymbol B_P^t
\boldsymbol A_P^t(\boldsymbol A_P^t)^\top\\
&=\boldsymbol B_P^t.
\end{align*}
Thus recalibration preserves the plastic coefficients exactly.

Since
$\boldsymbol A^t(\boldsymbol A^t)^\top=\boldsymbol I_r$,
multiplication by $\boldsymbol A^t$ preserves the
Frobenius norm of coefficient matrices.
Therefore,
\begin{align}
\|\Delta\boldsymbol W_{\mathrm{priv}}^t
  -\Delta\boldsymbol W^t\|_F^2
&=
\|\boldsymbol B^t-\widehat{\boldsymbol B}^t\|_F^2
\nonumber\\
&=
\sum_{j=1}^{r_S}c_j^t(1-\lambda_j^t)^2.
\label{eq:deviation}
\end{align}
Likewise,
\[
\|\Delta\boldsymbol W_{\mathrm{priv}}^t\|_F^2
=
\sum_{j=1}^{r_S}c_j^t+\|\boldsymbol B_P^t\|_F^2,
\qquad
\|\Delta\boldsymbol W^t\|_F^2
=
\sum_{j=1}^{r_S}c_j^t(\lambda_j^t)^2
+\|\boldsymbol B_P^t\|_F^2.
\]
Subtracting yields
\begin{equation}
\|\Delta\boldsymbol W_{\mathrm{priv}}^t\|_F^2
-\|\Delta\boldsymbol W^t\|_F^2
=
\sum_{j=1}^{r_S}
c_j^t\bigl(1-(\lambda_j^t)^2\bigr)
\ge0.
\label{eq:norm-reduction}
\end{equation}
Hence the update norm cannot increase.

\paragraph{(iii) Historical overlap reduction.}
Define
$\boldsymbol D_H^t
=\operatorname{diag}(h_1^t,\ldots,h_{r_S}^t)$.
Because the stable rows are eigenvectors of
$\boldsymbol H^{t-1}$,
\[
\boldsymbol A_S^t\boldsymbol H^{t-1}
=\boldsymbol D_H^t\boldsymbol A_S^t.
\]
Combining this identity with row orthonormality and
stable--plastic orthogonality gives
\[
\boldsymbol A_S^t\boldsymbol H^{t-1}
(\boldsymbol A_S^t)^\top
=\boldsymbol D_H^t,
\qquad
\boldsymbol A_S^t\boldsymbol H^{t-1}
(\boldsymbol A_P^t)^\top
=\boldsymbol0.
\]
The transposed cross term also vanishes because
$\boldsymbol H^{t-1}$ is symmetric.

Recall
$\mathcal I_t(\boldsymbol U)
=\operatorname{tr}
(\boldsymbol U\boldsymbol H^{t-1}\boldsymbol U^\top)$.
Expanding this expression for the two updates,
the vanishing cross terms imply
\begin{align*}
\mathcal I_t(\Delta\boldsymbol W_{\mathrm{priv}}^t)
&=
\operatorname{tr}\!\left(
\boldsymbol B_S^t\boldsymbol D_H^t
(\boldsymbol B_S^t)^\top
\right)
+\mathcal I_t(\boldsymbol B_P^t\boldsymbol A_P^t)\\
&=
\sum_{j=1}^{r_S}h_j^tc_j^t
+\mathcal I_t(\boldsymbol B_P^t\boldsymbol A_P^t),
\\[2pt]
\mathcal I_t(\Delta\boldsymbol W^t)
&=
\operatorname{tr}\!\left(
\boldsymbol B_S^t\boldsymbol\Lambda_S^t
\boldsymbol D_H^t\boldsymbol\Lambda_S^t
(\boldsymbol B_S^t)^\top
\right)
+\mathcal I_t(\boldsymbol B_P^t\boldsymbol A_P^t)\\
&=
\sum_{j=1}^{r_S}h_j^tc_j^t(\lambda_j^t)^2
+\mathcal I_t(\boldsymbol B_P^t\boldsymbol A_P^t).
\end{align*}
The plastic contribution is identical in both
expressions and cancels upon subtraction:
\begin{equation}
\mathcal I_t(\Delta\boldsymbol W_{\mathrm{priv}}^t)
-\mathcal I_t(\Delta\boldsymbol W^t)
=
\sum_{j=1}^{r_S}
h_j^tc_j^t\bigl(1-(\lambda_j^t)^2\bigr)
\ge0.
\label{eq:interference-reduction}
\end{equation}
Every summand is nonnegative since
$h_j^t,c_j^t\ge0$ and $0\le\lambda_j^t<1$.
The decrease is strict whenever
$h_j^tc_j^t>0$ for at least one stable direction.
This proves all claims of the theorem.
Empty branches are interpreted using empty matrices
and zero sums.
\hfill$\square$\par

\paragraph{Interpretation.}
These identities hold for each realized private update,
without assuming that $c_j^t$ is a noise-free measure
of task relevance.
They establish exact preservation of plastic
coefficients and non-increasing historical overlap;
their implications for old-task loss are addressed
separately in Theorem~\ref{thm:forgetting_main}.

\subsection{Proof of the Conditional Old-Task Loss Bound}
\label{app:forgetting}

\paragraph{Proof of Theorem~\ref{thm:forgetting_main}.}
Fix a task $t\ge2$, the pre-merge weight
$\boldsymbol W^{t-1}$, and the realized history and
private coefficients.
Consider the two updates
$\Delta\boldsymbol W_{\mathrm{priv}}^t$ and
$\Delta\boldsymbol W^t$ before and after recalibration.
Throughout the proof, both updates are evaluated from
the same pre-merge weight.

For matrices of the same dimension, write
$\langle\boldsymbol X,\boldsymbol Y\rangle_F
=\operatorname{tr}(\boldsymbol X^\top\boldsymbol Y)$
for the Frobenius inner product.
The notation
$\nabla^2F_{\mathrm{old}}(\boldsymbol W)
[\boldsymbol V,\boldsymbol V]$
denotes the Hessian quadratic form in direction
$\boldsymbol V$.
Recall
\[
\mathcal I_t(\boldsymbol V)
=
\operatorname{tr}\!\left(
\boldsymbol V\boldsymbol H^{t-1}\boldsymbol V^\top
\right).
\]

By the theorem's assumptions,
\[
\|\nabla F_{\mathrm{old}}(\boldsymbol W^{t-1})\|_F
\le\nu.
\]
Moreover, the same constants $\kappa,\rho\ge0$
satisfy
\[
\nabla^2F_{\mathrm{old}}
(\boldsymbol W^{t-1}+s\boldsymbol U)
[\boldsymbol V,\boldsymbol V]
\le
\kappa\mathcal I_t(\boldsymbol V)
+\rho\|\boldsymbol V\|_F^2
\]
for both
$\boldsymbol U\in
\{\Delta\boldsymbol W_{\mathrm{priv}}^t,
  \Delta\boldsymbol W^t\}$,
every $s\in[0,1]$, and every
$\boldsymbol V\in\mathbb R^{m\times n}$.
The common constants and history define the same
upper-bound function for both updates.

\paragraph{(i) Upper bound on the old-task loss increase.}
For either admissible update $\boldsymbol U$, define
the scalar function
\[
f_{\boldsymbol U}(s)
=
F_{\mathrm{old}}(\boldsymbol W^{t-1}+s\boldsymbol U),
\qquad s\in[0,1].
\]
The differentiability assumption gives
\[
f_{\boldsymbol U}'(0)
=
\langle
\nabla F_{\mathrm{old}}(\boldsymbol W^{t-1}),
\boldsymbol U
\rangle_F,
\qquad
f_{\boldsymbol U}''(s)
=
\nabla^2F_{\mathrm{old}}
(\boldsymbol W^{t-1}+s\boldsymbol U)
[\boldsymbol U,\boldsymbol U].
\]
Taylor's formula with integral remainder yields
\begin{align*}
&F_{\mathrm{old}}(\boldsymbol W^{t-1}+\boldsymbol U)
-F_{\mathrm{old}}(\boldsymbol W^{t-1})\\
&\quad=
\langle
\nabla F_{\mathrm{old}}(\boldsymbol W^{t-1}),
\boldsymbol U
\rangle_F\\
&\qquad\quad+
\int_0^1(1-s)
\nabla^2F_{\mathrm{old}}
(\boldsymbol W^{t-1}+s\boldsymbol U)
[\boldsymbol U,\boldsymbol U]\,ds.
\end{align*}
By Cauchy--Schwarz and the gradient bound,
\[
\langle
\nabla F_{\mathrm{old}}(\boldsymbol W^{t-1}),
\boldsymbol U
\rangle_F
\le
\|\nabla F_{\mathrm{old}}(\boldsymbol W^{t-1})\|_F
\|\boldsymbol U\|_F
\le\nu\|\boldsymbol U\|_F.
\]
Applying the curvature bound with
$\boldsymbol V=\boldsymbol U$ and using
$\int_0^1(1-s)\,ds=1/2$, we obtain
\begin{align*}
&\int_0^1(1-s)
\nabla^2F_{\mathrm{old}}
(\boldsymbol W^{t-1}+s\boldsymbol U)
[\boldsymbol U,\boldsymbol U]\,ds\\
&\qquad\le
\frac{\kappa}{2}\mathcal I_t(\boldsymbol U)
+\frac{\rho}{2}\|\boldsymbol U\|_F^2.
\end{align*}
Combining these inequalities proves
\[
F_{\mathrm{old}}(\boldsymbol W^{t-1}+\boldsymbol U)
-F_{\mathrm{old}}(\boldsymbol W^{t-1})
\le\mathcal U_t(\boldsymbol U),
\]
where
\[
\mathcal U_t(\boldsymbol U)
=
\nu\|\boldsymbol U\|_F
+\frac{\kappa}{2}\mathcal I_t(\boldsymbol U)
+\frac{\rho}{2}\|\boldsymbol U\|_F^2.
\]
This inequality holds for both updates.

\paragraph{(ii) Comparison before and after recalibration.}
Equations~\ref{eq:norm-reduction}
and~\ref{eq:interference-reduction} give
\begin{align*}
\|\Delta\boldsymbol W_{\mathrm{priv}}^t\|_F^2
-\|\Delta\boldsymbol W^t\|_F^2
&=
\sum_{j=1}^{r_S}
c_j^t\bigl(1-(\lambda_j^t)^2\bigr)\ge0,\\
\mathcal I_t(\Delta\boldsymbol W_{\mathrm{priv}}^t)
-\mathcal I_t(\Delta\boldsymbol W^t)
&=
\sum_{j=1}^{r_S}
h_j^tc_j^t\bigl(1-(\lambda_j^t)^2\bigr)\ge0.
\end{align*}
Since norms are nonnegative, the first inequality also
implies
\[
\|\Delta\boldsymbol W^t\|_F
\le\|\Delta\boldsymbol W_{\mathrm{priv}}^t\|_F.
\]
Subtracting the two upper bounds and substituting
the identities above yields
\begin{align}
&\mathcal U_t(\Delta\boldsymbol W_{\mathrm{priv}}^t)
-\mathcal U_t(\Delta\boldsymbol W^t)
\nonumber\\
&\quad=
\nu\left(
\|\Delta\boldsymbol W_{\mathrm{priv}}^t\|_F
-\|\Delta\boldsymbol W^t\|_F
\right)
\nonumber\\
&\qquad\quad+
\frac12\sum_{j=1}^{r_S}
(\kappa h_j^t+\rho)c_j^t
\bigl(1-(\lambda_j^t)^2\bigr)
\ge0.
\label{eq:certificate-gap}
\end{align}
Every term is nonnegative because
$\nu,\kappa,\rho,h_j^t,c_j^t\ge0$
and $0\le\lambda_j^t<1$.
Therefore,
\[
\mathcal U_t(\Delta\boldsymbol W^t)
\le
\mathcal U_t(\Delta\boldsymbol W_{\mathrm{priv}}^t),
\]
which proves the theorem.
\hfill$\square$\par

\paragraph{Interpretation.}
The theorem compares a common upper bound for two
updates at the same pre-merge model.
It does not imply an ordering of their actual
old-task losses or of complete training trajectories
with different histories.
The connection between historical overlap and
old-task loss enters through the stated curvature
assumption.

\paragraph{Extension to multiple adapted layers.}
Let $\ell=1,\ldots,L_{\mathrm{ad}}$ index the adapted
weight matrices, and collect their increments as
$\boldsymbol U=(\boldsymbol U_\ell)_{\ell=1}^{L_{\mathrm{ad}}}$.
On this product space, define
\[
\|\boldsymbol U\|_{\mathrm{joint}}^2
=
\sum_{\ell=1}^{L_{\mathrm{ad}}}
\|\boldsymbol U_\ell\|_F^2,
\qquad
\mathcal I_t^{\mathrm{joint}}(\boldsymbol U)
=
\sum_{\ell=1}^{L_{\mathrm{ad}}}
\operatorname{tr}\!\left(
\boldsymbol U_\ell\boldsymbol H_\ell^{t-1}
\boldsymbol U_\ell^\top
\right).
\]
Here $\boldsymbol H_\ell^{t-1}$ is the history matrix
for layer $\ell$.
Let $\boldsymbol W^{t-1}$ now denote the collection
of pre-merge matrices, with all other model parameters
held fixed in the comparison.

Suppose the old-task loss is twice continuously
differentiable along both joint update segments,
its joint gradient satisfies
\[
\|\nabla F_{\mathrm{old}}
(\boldsymbol W^{t-1})\|_{\mathrm{joint}}
\le\nu,
\]
and its full Hessian satisfies
\[
\nabla^2F_{\mathrm{old}}
(\boldsymbol W^{t-1}+s\boldsymbol U)
[\boldsymbol V,\boldsymbol V]
\le
\kappa\mathcal I_t^{\mathrm{joint}}(\boldsymbol V)
+\rho\|\boldsymbol V\|_{\mathrm{joint}}^2
\]
for both joint updates, all $s\in[0,1]$, and every
joint perturbation $\boldsymbol V$.
This is a bound on the full Hessian, including
cross-layer terms, rather than separate bounds
on individual layers.

The same Taylor argument gives the joint upper bound
\[
\mathcal U_t^{\mathrm{joint}}(\boldsymbol U)
=
\nu\|\boldsymbol U\|_{\mathrm{joint}}
+\frac{\kappa}{2}
\mathcal I_t^{\mathrm{joint}}(\boldsymbol U)
+\frac{\rho}{2}\|\boldsymbol U\|_{\mathrm{joint}}^2.
\]
Applying the recalibration identities to each layer
and summing shows that neither the joint squared
norm nor the joint historical overlap increases.
Hence
\[
\mathcal U_t^{\mathrm{joint}}(\Delta\boldsymbol W^t)
\le
\mathcal U_t^{\mathrm{joint}}
(\Delta\boldsymbol W_{\mathrm{priv}}^t).
\]
Thus the single-matrix argument extends to multiple
adapted layers under the stated joint assumptions.

\section{Supplementary Analysis of the Bilinear Noise Interaction}
\label{app:noise}

This section supplements Observation~2 and
Section~\ref{sec:private_optimization} by quantifying
the fresh-noise energy in one DP-SGD step.

\paragraph{Setup.}
Condition on the optimization state and current minibatch
before drawing fresh noise. Let
$\bar{\boldsymbol B}\in\mathbb R^{m\times r}$ and
$\bar{\boldsymbol A}\in\mathbb R^{r\times n}$
be the factors after the clipped-gradient update without
this step's noise.
Let $\boldsymbol N_B\in\mathbb R^{m\times r}$ and
$\boldsymbol N_A\in\mathbb R^{r\times n}$ have mutually
independent centered Gaussian entries with conditional
variances $v_B^2$ and $v_A^2$, respectively.
All expectations below are over these fresh perturbations,
conditional on the fixed state and minibatch.

\begin{proposition}[Single-step noise energy]
\label{prop:noise}
Under the preceding assumptions,
\begin{align}
&\mathbb E\left\|
(\bar{\boldsymbol B}+\boldsymbol N_B)
(\bar{\boldsymbol A}+\boldsymbol N_A)
-\bar{\boldsymbol B}\bar{\boldsymbol A}
\right\|_F^2
\nonumber\\
&\qquad=
n v_A^2\|\bar{\boldsymbol B}\|_F^2
+m v_B^2\|\bar{\boldsymbol A}\|_F^2
+mnr\,v_A^2v_B^2.
\label{eq:jointnoise}
\end{align}
For a fixed basis $\boldsymbol A^t$ satisfying
$\boldsymbol A^t(\boldsymbol A^t)^\top=\boldsymbol I_r$,
\begin{equation}
\|\boldsymbol N_B\boldsymbol A^t\|_F^2
=\|\boldsymbol N_B\|_F^2,
\qquad
\mathbb E\|\boldsymbol N_B\boldsymbol A^t\|_F^2
=mr\,v_B^2.
\label{eq:single-noise}
\end{equation}
\end{proposition}

\paragraph{Proof of Proposition~\ref{prop:noise}.}
The joint perturbation expands as
$\bar{\boldsymbol B}\boldsymbol N_A
+\boldsymbol N_B\bar{\boldsymbol A}
+\boldsymbol N_B\boldsymbol N_A$.
All pairwise Frobenius inner products have zero expectation:
the first two terms are independently centered;
for the first and third, condition additionally on
$\boldsymbol N_A$ and average over $\boldsymbol N_B$;
for the second and third, reverse this conditioning.
Thus the expected squared norm is the sum of the
three expected squared norms.

Using
$\mathbb E[\boldsymbol N_A\boldsymbol N_A^\top]
=n v_A^2\boldsymbol I_r$ and
$\mathbb E[\boldsymbol N_B^\top\boldsymbol N_B]
=m v_B^2\boldsymbol I_r$ gives
\begin{align*}
\mathbb E\|\bar{\boldsymbol B}\boldsymbol N_A\|_F^2
&=n v_A^2\|\bar{\boldsymbol B}\|_F^2,\\
\mathbb E\|\boldsymbol N_B\bar{\boldsymbol A}\|_F^2
&=m v_B^2\|\bar{\boldsymbol A}\|_F^2,\\
\mathbb E\|\boldsymbol N_B\boldsymbol N_A\|_F^2
&=n v_A^2\mathbb E\|\boldsymbol N_B\|_F^2
=mnr\,v_A^2v_B^2.
\end{align*}
The last equality follows by first conditioning on
$\boldsymbol N_B$. Summing proves
Eq.~\ref{eq:jointnoise}.
For the fixed-basis case,
$\|\boldsymbol N_B\boldsymbol A^t\|_F^2
=\operatorname{tr}(
\boldsymbol N_B\boldsymbol A^t
(\boldsymbol A^t)^\top\boldsymbol N_B^\top)
=\|\boldsymbol N_B\|_F^2$.
Taking expectations proves Eq.~\ref{eq:single-noise}.
\hfill$\square$\par

\paragraph{Interpretation and scope.}
The term $mnr\,v_A^2v_B^2$ is the expected energy
of the bilinear interaction in \textbf{Observation~2}.
Fixing $\boldsymbol A^t$ removes both perturbation
terms involving $\boldsymbol N_A$, while its row
orthonormality preserves the remaining coefficient-noise
energy. At $\bar{\boldsymbol A}=\boldsymbol A^t$
and the same $v_B$, the joint-factor expected energy
exceeds the single-factor value by the nonnegative
quantity
$n v_A^2\|\bar{\boldsymbol B}\|_F^2+mnr\,v_A^2v_B^2$.
The empirical measurements in Observation~2 assess
the interaction's relative magnitude during training.

For SGD with learning rate $\eta$, clipping norm $C$,
noise multiplier $\sigma$, and normalization $b>0$
fixed under the above conditioning, the parameter-noise
standard deviation is $v=\eta\sigma C/b$.
Adaptive optimizers require a separate analysis of
how gradient noise is transformed.
These results concern conditional single-step
perturbations, not final model errors or accuracies:
clipped gradients and subsequent trajectories may differ
between joint-factor and single-factor training.
They also do not characterize noise after nonlinear,
data-dependent recalibration.

\section{Supplementary Experimental Information}
\label{app:experimental}

\subsection{Experimental details}
\label{app:setup}
\paragraph{Data splits and task construction.}
We follow the training/test splits and class-incremental task construction of \citet{he2026task} for the five benchmarks used in our experiments.
We evaluate ImageNet-R with $T\in\{5,10,20\}$,
ImageNet-A with $T\in\{10,20\}$, and CIFAR-100,
CUB, and DomainNet with $T=10$.
Task class sets are disjoint. After each task, we
evaluate the model over all classes observed so far,
without access to task identity. 

\paragraph{Privacy budget.}
Each disjoint task partition is assigned the same
privacy budget $(\epsilon,\delta)$.
Provided that all private-data-dependent operations
within each task are covered by this budget and
subsequent tasks access earlier partitions only
through DP outputs, parallel composition yields
the same $(\epsilon,\delta)$ guarantee for the
entire continual learning process.

\paragraph{Baseline selection.}
We use AB-LoRA and FA-LoRA as controlled baselines,
both with rank $r=32$, matching the total rank of
\textsf{DP-SPA}.
AB-LoRA trains both low-rank factors, whereas
FA-LoRA fixes the input-side factor and trains
only the output-side factor.
This comparison examines whether the benefits of
\textsf{DP-SPA} extend beyond fixing one factor
under differential privacy.

Our comparison focuses on LoRA-based adaptation
under DP rather than a comprehensive evaluation
of continual learning methods.
Non-private continual learning results are not
directly comparable at a fixed privacy budget.
Adapting methods with additional private-data-dependent
operations requires accounting for those operations
as well as gradient updates, and may require
algorithmic changes.
We therefore restrict our controlled comparison
to these two baselines.

\subsection{Additional Results}
\label{app:Additional Results}

\paragraph{Comparison of recalibration strategies.}
To assess whether direction-wise recalibration provides benefits
beyond uniform shrinkage, we compare four strategies:
(i) no recalibration;
(ii) constant scaling of the stable branch by $\gamma=0.5$;
(iii) norm-matched uniform scaling; and
(iv) the proposed history-aware recalibration.
All variants use the same stable--plastic basis construction
and private optimization protocol, with the plastic branch
retained without rescaling.

For norm-matched scaling, each LoRA module computes
$\lambda_j^t$ from its own history and private update,
but applies a uniform coefficient
\begin{equation}
    \gamma^t =
    \sqrt{\frac{\sum_{j=1}^{r_S}c_j^t(\lambda_j^t)^2}
    {\sum_{j=1}^{r_S}c_j^t}},
\end{equation}
with $\gamma^t=0$ when the denominator is zero.
The row orthonormality of $\boldsymbol{A}_S^t$ ensures
$\|\gamma^t\boldsymbol{B}_S^t\boldsymbol{A}_S^t\|_F
=\|\boldsymbol{B}_S^t\boldsymbol{\Lambda}_S^t
\boldsymbol{A}_S^t\|_F$.
This matches the stable-branch update norm at the same
pre-merge state while removing direction-wise scaling.
Each variant runs the full continual sequence, maintains
its own history using its merged updates, and omits
recalibration on the first task.

Table~\ref{tab:recalibration_strategies} reports the results
on ImageNet-R and ImageNet-A with $T=20$.
DP-SPA achieves the highest mean last and average accuracies
on both datasets. Compared with no recalibration, it improves
$\mathcal{A}_{Last}$ by 0.89 and 0.88 percentage points,
respectively. Constant scaling underperforms no recalibration
on both datasets, indicating that a fixed shrinkage factor
does not reliably improve performance.
Norm-matched scaling yields modest gains in
$\mathcal{A}_{Last}$ but does not consistently improve
$\mathcal{A}_{Avg}$.
DP-SPA further exceeds norm-matched scaling by 0.57 and 0.64
percentage points in $\mathcal{A}_{Last}$, and by 0.69 and
1.57 points in $\mathcal{A}_{Avg}$, respectively.

\begin{table*}[h]
    \centering
    \caption{Comparison of recalibration strategies on ImageNet-R and
    ImageNet-A with $T=20$.
    Constant scaling uses $\gamma=0.5$, while norm-matched scaling
    matches the stable-branch update norm of direction-wise
    recalibration at the same pre-merge state.}
    \vspace{-1mm}
    \label{tab:recalibration_strategies}
    \small
    \setlength{\tabcolsep}{4pt}
    \renewcommand{\arraystretch}{1.10}
    \setlength{\heavyrulewidth}{1.0pt}
    \setlength{\lightrulewidth}{0.5pt}
    \setlength{\cmidrulewidth}{0.4pt}
    \begin{tabular}{lcccc}
        \toprule
        \multirow{2}{*}{\textbf{Strategy}}
        & \multicolumn{2}{c}{\textbf{ImageNet-R} ($T{=}20$)}
        & \multicolumn{2}{c}{\textbf{ImageNet-A} ($T{=}20$)} \\

        \cmidrule(lr){2-3}
        \cmidrule(lr){4-5}

        & $\mathcal{A}_{Last}\uparrow$
        & $\mathcal{A}_{Avg}\uparrow$
        & $\mathcal{A}_{Last}\uparrow$
        & $\mathcal{A}_{Avg}\uparrow$ \\
        \midrule

        No recalibration
        & $67.25_{\pm 0.69}$ & $77.49_{\pm 0.72}$
        & $47.56_{\pm 0.63}$ & $59.68_{\pm 0.55}$ \\

        Constant scaling
        & $65.30_{\pm 0.54}$ & $76.83_{\pm 0.61}$
        & $44.97_{\pm 0.67}$ & $58.18_{\pm 0.52}$\\

        Norm-matched scaling
        & $67.57_{\pm 0.59}$ & $77.65_{\pm 0.42}$
        & $47.80_{\pm 0.49}$ & $59.32_{\pm 0.60}$ \\

        \textbf{DP-SPA}
        & $68.14_{\pm 0.75}$ & $78.34_{\pm 0.67}$
        & $48.44_{\pm 0.84}$ & $60.89_{\pm 0.78}$ \\
        \bottomrule
    \end{tabular}
\end{table*}

\paragraph{Sensitivity to stable--plastic rank allocation.}
Table~\ref{tab:rank_allocation} compares different allocations
with fixed total rank $r=32$, $T=20$, and $\epsilon=3$.
The best mean accuracies are obtained with $(r_S,r_P)=(16,16)$
on ImageNet-R and $(8,24)$ on ImageNet-A, indicating
dataset-dependent allocation preferences.
All evaluated mixed allocations outperform both single-branch
configurations on both metrics, supporting the complementary
roles of stable and plastic directions.
We use $(16,16)$ as a common default across datasets.

\begin{table*}[h]
    \centering
    \caption{Sensitivity to stable--plastic rank allocation on
    ImageNet-R and ImageNet-A with $T=20$, $\epsilon=3$,
    and fixed total rank $r_S+r_P=32$.}
    \vspace{-1mm}
    \label{tab:rank_allocation}
    \small
    \setlength{\tabcolsep}{5pt}
    \renewcommand{\arraystretch}{1.10}
    \setlength{\heavyrulewidth}{1.0pt}
    \setlength{\lightrulewidth}{0.5pt}
    \setlength{\cmidrulewidth}{0.4pt}
    \begin{tabular}{lcccc}
        \toprule
        \multirow{2}{*}{$(r_S,r_P)$}
        & \multicolumn{2}{c}{\textbf{ImageNet-R} ($T{=}20$)}
        & \multicolumn{2}{c}{\textbf{ImageNet-A} ($T{=}20$)} \\

        \cmidrule(lr){2-3}
        \cmidrule(lr){4-5}

        & $\mathcal{A}_{Last}\uparrow$
        & $\mathcal{A}_{Avg}\uparrow$
        & $\mathcal{A}_{Last}\uparrow$
        & $\mathcal{A}_{Avg}\uparrow$ \\
        \midrule

        $(0,32)$
        & $65.80_{\pm 0.74}$ & $76.32_{\pm 0.81}$
        & $45.19_{\pm 0.90}$ & $58.24_{\pm 0.68}$ \\

        $(8,24)$
        & $67.91_{\pm 0.82}$ & $77.63_{\pm 0.78}$
        & $49.12_{\pm 0.95}$ & $61.06_{\pm 0.77}$ \\

        $(16,16)$ 
        & $68.14_{\pm 0.75}$ & $78.34_{\pm 0.67}$
        & $48.44_{\pm 0.84}$ & $60.89_{\pm 0.78}$ \\

        $(24,8)$
        & $67.93_{\pm 0.30}$ & $77.83_{\pm 0.28}$
        & $47.56_{\pm 0.46}$ & $59.97_{\pm 0.34}$ \\

        $(32,0)$
        & $67.76_{\pm 0.60}$ & $77.15_{\pm 0.46}$
        & $47.05_{\pm 0.65}$ & $59.43_{\pm 0.52}$ \\
        \bottomrule
    \end{tabular}
\end{table*}

\begin{figure}[ht]
\centering
\begin{subfigure}[t]{0.33\textwidth}
\centering
\includegraphics[width=\linewidth]{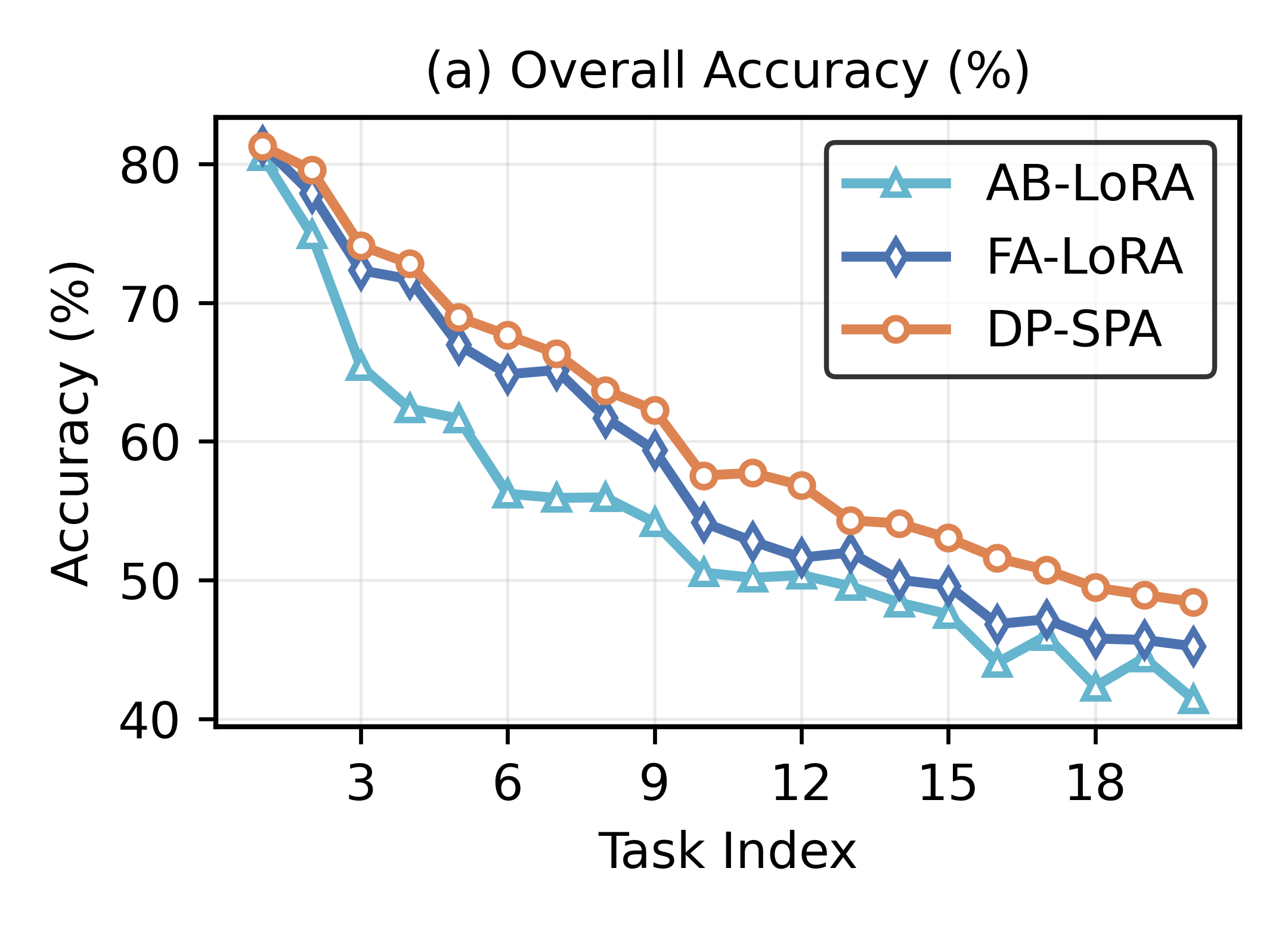}
\end{subfigure}\hfill
\begin{subfigure}[t]{0.33\textwidth}
\centering
\includegraphics[width=\linewidth]{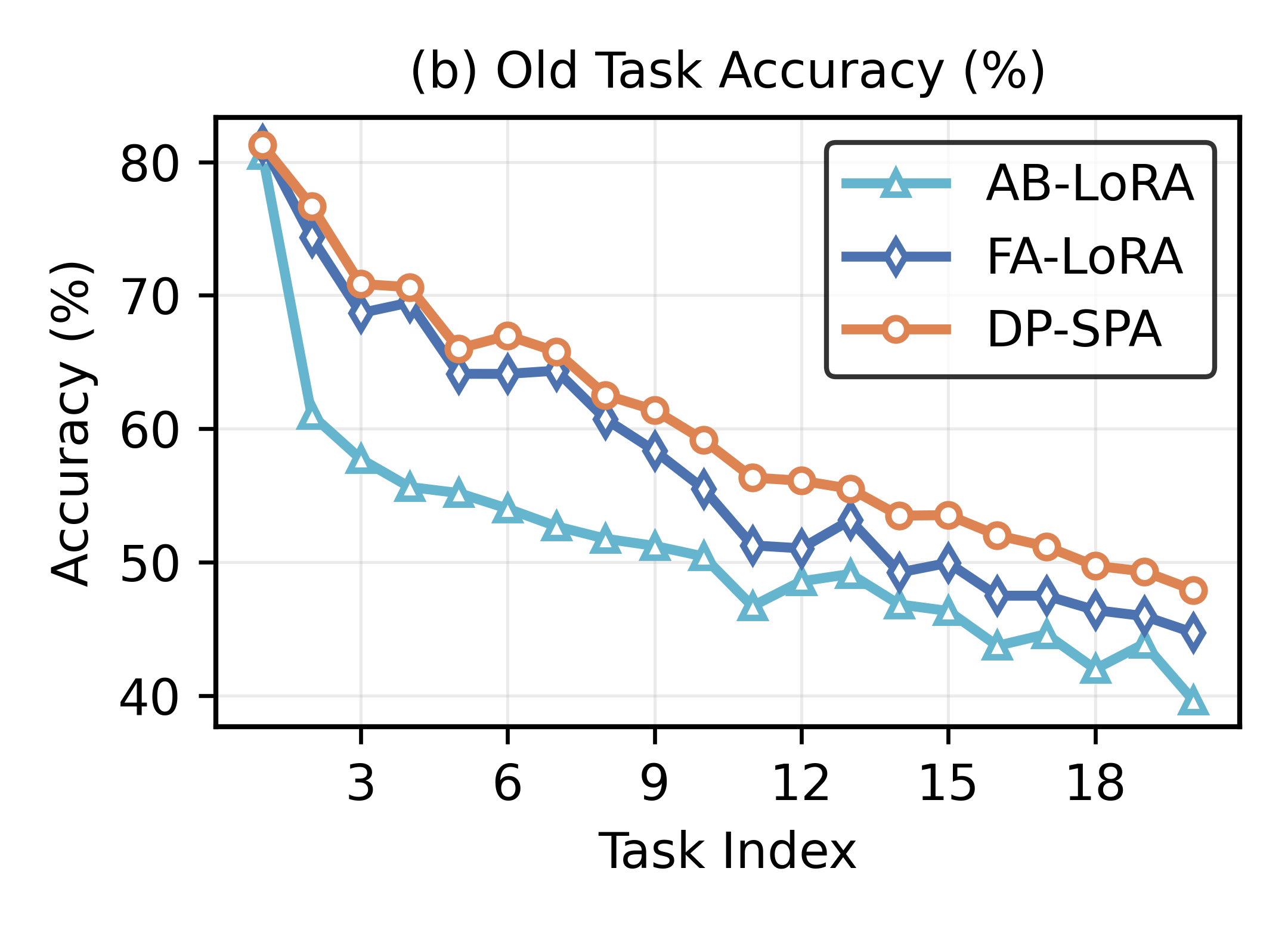}
\end{subfigure}\hfill
\begin{subfigure}[t]{0.33\textwidth}
\centering
\includegraphics[width=\linewidth]{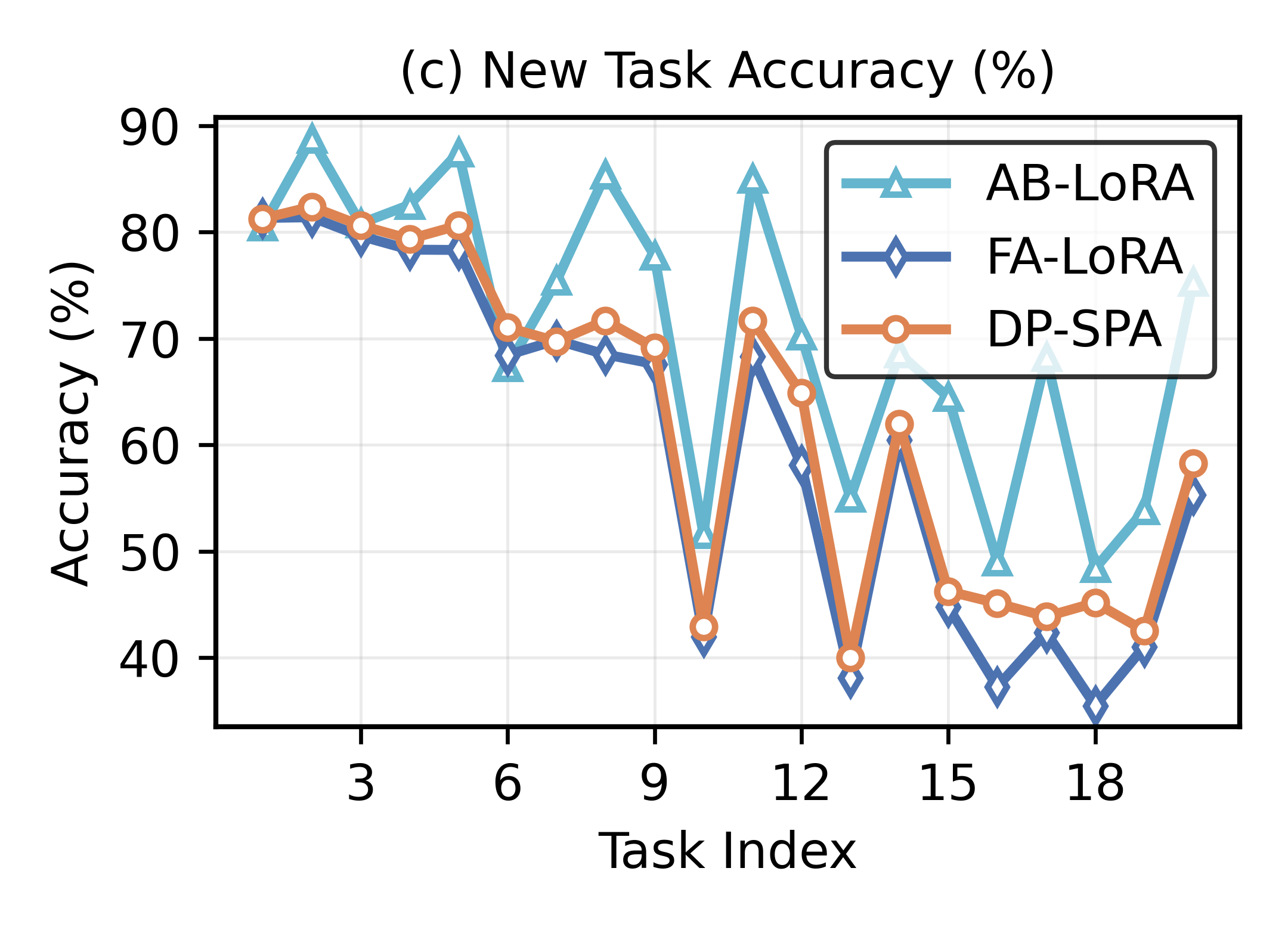}
\end{subfigure}
\caption{Overall, old-task, and new-task accuracy across 20 incremental tasks
on ImageNet-A.}
\label{fig:imagenet_a}
\end{figure}

\paragraph{Performance on ImageNet-A.}
Figure~\ref{fig:imagenet_a} compares overall, old-task, and new-task accuracy across 20 incremental tasks on ImageNet-A.
DP-SPA achieves higher overall and old-task accuracy through most stages while maintaining new-task accuracy comparable to FA-LoRA, suggesting that improved retention is the main contributor to its overall advantage.

\end{document}